\documentclass[10pt,twocolumn,letterpaper]{article}

\usepackage{cvpr}              

\usepackage{multirow}
\usepackage{booktabs}
\usepackage[dvipsnames]{xcolor}
\usepackage[export]{adjustbox}
\usepackage{xurl}
\usepackage[accsupp]{axessibility}

\newcommand{\ours}{\textbf{\textsc{SelVA}}\xspace}
\newcommand{\ourbench}{{\textsc{VGG-MonoAudio}}\xspace}
\newcommand{\reg}{\texttt{[SUP]}\xspace}
\newcommand{\ba}{\mathbf{a}}
\newcommand{\bt}{\mathbf{t}}
\newcommand{\bv}{\mathbf{v}}
\newcommand{\bc}{\mathbf{c}}
\newcommand{\tar}{\text{tar}}
\newcommand{\pair}{\text{pair}}

\newcommand{\hz}{\vphantom{\parbox[c]{0.25cm}{\rule{0.25cm}{0.28cm}}}}
\newcommand{\learn}[1]{{\setlength\fboxsep{1.5pt}\colorbox{red!20}{\hz{$\displaystyle #1$}}}}
\newcommand{\frozen}[1]{{\setlength\fboxsep{1.5pt}\colorbox{blue!20}{\hz{$\displaystyle #1$}}}}

\newcommand{\fire}{%
   \includegraphics[height=1em, valign=m]{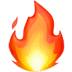}%
}

\newcommand{\snowflake}{%
   \includegraphics[height=1em, valign=m]{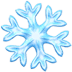}%
}
\newcommand{\demourl}{\texttt{\url{https://jnwnlee.github.io/selva-demo/}}}

\usepackage{colortbl}
\definecolor{backBlue}{RGB}{230, 244, 252}
\definecolor{backGray}{RGB}{230, 230, 230}
\definecolor{backRed}{RGB}{252, 225, 230}

\renewcommand{\paragraph}[1]{\noindent\textbf{#1.}}

\newcommand\blfootnote[1]{%
  \begingroup
  \renewcommand\thefootnote{}\footnote{#1}%
  \addtocounter{footnote}{-1}%
  \endgroup
}

\definecolor{cvprblue}{rgb}{0.21,0.49,0.74}
\usepackage[pagebackref,breaklinks,colorlinks,allcolors=cvprblue]{hyperref}

\def\paperID{22781} 
\def\confName{CVPR}
\def\confYear{2026}

\title{Hear What Matters! Text-conditioned Selective Video-to-Audio Generation}

\author{
Junwon Lee$^{1,\dagger}$ \qquad
Juhan Nam\textsuperscript{1,2} \qquad
Jiyoung Lee$^{3*}$\vspace{.5em}
\\ 
\textsuperscript{1}Graduate School of AI and \textsuperscript{2}Graduate School of Cultural Technology, KAIST\\
\textsuperscript{3}Division of AI and Software,
Ewha Womans University\\
{\tt\small 
    \{james39,juhan.nam\}@kaist.ac.kr, lee.jiyoung@ewha.ac.kr
}\vspace{.5em}\\
{\demourl}
}

\begin{document}
\maketitle
\begin{abstract}
This work introduces a new task, text-conditioned selective video-to-audio (V2A) generation, which produces only the user-intended sound from a multi-object video.
This capability is especially crucial in multimedia production, where audio tracks are handled individually for each sound source for precise editing, mixing, and creative control.
We propose \ours, a novel text-conditioned V2A model that treats the text prompt as an explicit selector to distinctly extract prompt-relevant sound-source visual features from the video encoder. 
To suppress text-irrelevant activations with efficient video encoder finetuning, the proposed supplementary tokens promote cross-attention to yield robust semantic and temporal grounding.
\ours further employs an autonomous video-mixing scheme in a self-supervised manner to overcome the lack of mono audio track supervision. 
We evaluate \ours on \ourbench, a curated benchmark of clean single-source videos for such a task. 
Extensive experiments and ablations consistently verify its effectiveness across audio quality, semantic alignment, and temporal synchronization.
\blfootnote{
\hspace{-2em} $^{\dagger}$ Work partly done during an internship at NAVER AI Lab.\\
$^{*}$ Corresponding author.}
\end{abstract}    
\section{Introduction}
In a bustling café, you can effortlessly tune into a friend’s laughter amid the chatter, or pick out the sound of a violin from an entire orchestra.
This effortless segregation of sounds, achieved through \textit{auditory scene analysis}, is a hallmark of human perception~\cite{audio_scene_analysis}.
At the core of this process lies selective attention, which enables us to focus on a specific sound source while filtering out irrelevant noise.
Such an attention-driven mechanism allows humans to extract what truly matters from a rich and noisy world.

Recent advances in neural audio generation have enabled realistic sound synthesis from text descriptions or visual scenes in film and game post-production, known as Foley~\cite{oh2023demand}.
Video-to-audio (V2A) models~\cite{rewas,diff-foley,video-foley,mmaudio,multifoley} now generate temporally coherent, context-aware audio directly from visual content.
However, they typically produce a single holistic soundtrack at a time.
This limitation stands in sharp contrast to the sound production~\cite{oh2023demand,challengetta}, where sound designers do not sonify every visible object.
Instead, they build scenes by layering individually crafted tracks, selectively including or suppressing sound elements, and then mixing and mastering them, to achieve precise control.
However, with current V2A models, even minor omissions in the output require re-synthesizing the entire audio, hindering practical usability.

\begin{figure}[t!]
  \centering
  \includegraphics[width=\linewidth]{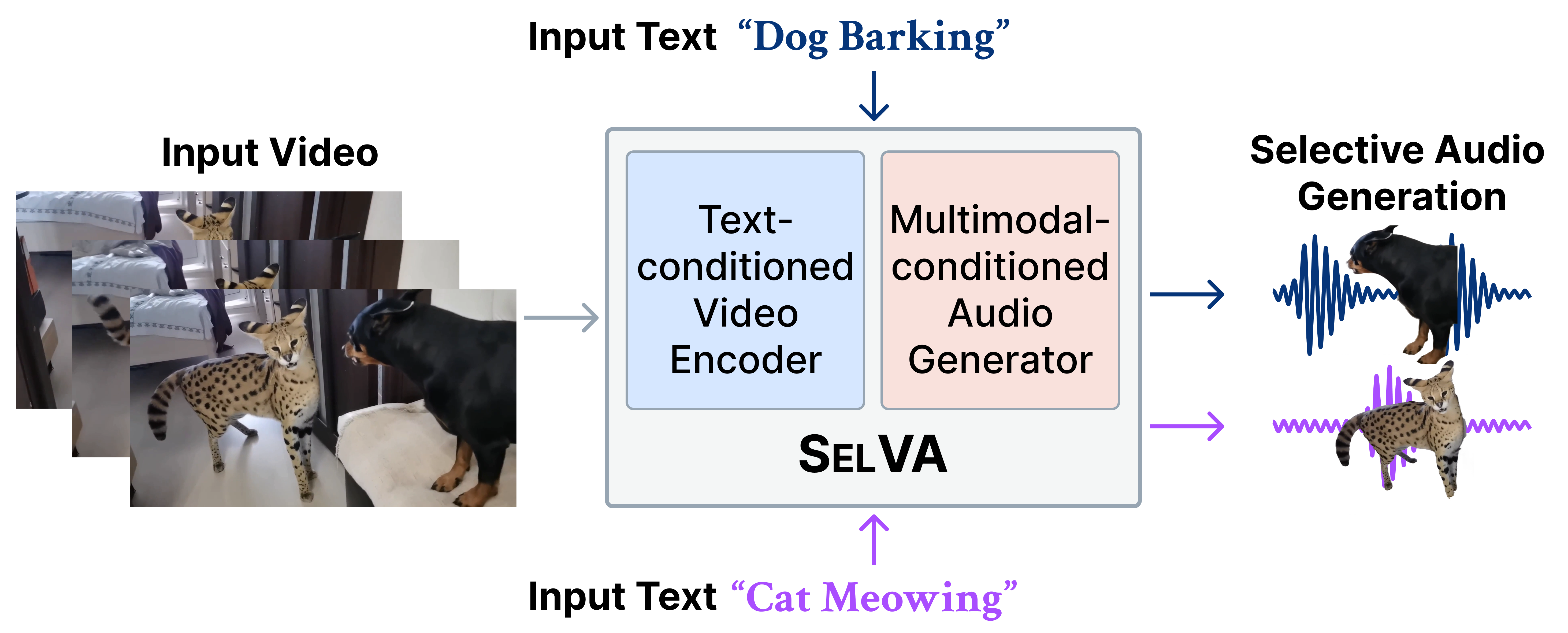}
  \vspace{-2em}
  \caption{\ours turns text prompts into precise selectors of sound sources within a video. The intent-focused video feature conditions the generator to synthesize only the user-specified sound source (\eg, `cat meowing' vs. `dog barking').}
   \label{fig:task}
\end{figure}

In this work, we tackle the \textbf{selective sound generation} problem: generating only the target sound that aligns with a user’s intention, given multimodal cues such as video and text.
However, the presence of multiple co-occurring sound sources in real-world videos makes it inherently difficult to isolate the target sound without explicit supervision. 
A few works~\cite{saganet,hearyourclick} have attempted to localize sound sources using off-the-shelf segmentation models, where visual prompts such as points or masks are used to specify the target object.
Yet, such approaches inherently operate on isolated spatial regions. 
Therefore, they struggle to handle broader visual context, including environmental or diffuse sounds (\eg, rainfall or wind), which cannot be localized to discrete visual boundaries.
In addition, reliance on large segmentation networks increases substantial computational overhead.
In contrast, we propose a new formulation of text-conditioned selective video-to-audio generation, where the target sound source is specified purely through a text prompt.
This removes the dependency on explicit spatial prompts while enabling more flexible and semantically expressive control over sound generation.

We propose a novel \ours method, consisting of two main modules: (1) a text-conditioned video encoder, and (2) a multimodal-conditioned selective audio generator.
Unlike previous works~\cite{mmaudio,rewas} that freeze visual-specific encoders, \ours efficiently trains the video encoder to use text prompts as explicit selectors of audible semantics.
To achieve a precise cross-modal grounding, we implement a cross-attention block in the video encoder with the proposed learnable token \reg. 
Inspired by the selective attention mechanism in human perception, \reg mitigates high-norm artifacts ~\cite{vit_register,cavmae_sync} in which models tend to highlight irrelevant tokens in attention blocks due to the spurious correlation.
To train \ours without explicitly source-separated groundtruth audios, we further employ an autonomous video-mixing strategy in a self-supervised manner, where two videos are spatially concatenated as input and the audio from one of them is used as the target.

To assess performance on such a novel task, we introduce \ourbench, a new evaluation benchmark comprising videos with a clear visual sound source corresponding to a mono audio track.
Experimental results demonstrate that \ours achieves state-of-the-art (SoTA) performance on \ourbench, showing robustness in terms of audio quality, semantic alignment, and temporal alignment.
Our contributions are summarized:
\begin{itemize}
    \item \ours is the first text-conditioned selective video-to-audio framework that relies solely on text prompts, without spatial guidance.
    \item A learnable supplementary token mitigates spurious cross-modal correlations while empowering intent-relevant cues, and the proposed video-mixing scheme enables selective learning without costly supervision.
    \item Experiments on new \ourbench demonstrate the effectiveness of \ours, showing superior audio quality, semantic fidelity, and temporal synchronization performance over existing SoTAs.
\end{itemize}
\section{Related Work}

\textbf{Cross-modal neural audio generation} has been widely explored due to its applicability in multimedia content production. 
Text-to-audio (T2A) aims to generate audio from an input text prompt, which usually describes the global semantics such as sound sources and their nuanced timbre (\eg, `drill buzzing harshly'). 
The common baseline is first to extract a text embedding from a pretrained text encoder such as CLAP~\cite{clap_ms,clap_laion} and T5~\cite{flant5}, then use it as a condition of generative models, including diffusion~\cite{audioldm,make-an-audio,stableaudio}, auto-regressive modeling~\cite{audiogen}, and flow-matching~\cite{tangoflux}. 
While text prompts offer intuitive semantic control, they inherently lack the ability to convey temporal dynamics of intensity or harmonics in audio~\cite{t-foley,fusionnet,sketch2sound}. 
Meanwhile, video-to-audio (V2A)~\cite{mmaudio,moviegen,multifoley,diff-foley,condfoleygen,specvqgan} resolves this issue by generating audio in synchrony with video. 
Such synchronization entails two complementary goals: semantic and temporal alignment. 
As a spatiotemporal modality, video conveys rich cues about sounding objects, including appearance, spatial location, and dynamic motion. 
In practice, current V2A frameworks remain strongly dependent on pretrained visual encoders~\cite{clip,tsn,vivit,diff-foley,synchformer} to provide the conditioning representations.

Recently, some works~\cite{sonicvisionlm,video-foley,rewas} have leveraged the capability of pretrained T2A models for V2A generation to reduce the training cost and ensure controllability. 
Most approaches~\cite{vatt,mmaudio,moviegen,audiogen-omni} naively hypothesize that complementary relations of video and text conditions, producing high-fidelity audio. 
Text prompts complement video embeddings by supplying semantic cues that visual encoders often miss (\eg, visual ambiguity such as occlusion of sounding objects caused by camera work)~\cite{v2a-occlusion,vatt}. 
For example, ReWaS~\cite{rewas} and Video-Foley~\cite{video-foley} rely on text to control the semantics of sound, while Multifoley~\cite{multifoley} leveraged text to change the sound timbre.
VinTAGe~\cite{vintage} generates both on-screen sound from visual cues and off-screen sound from textual cues.
However, existing works do not use text to specify \textit{which} sound sources should be heard. 
Instead, text serves merely as an auxiliary cue, not to modulate the given visual information. 
In this paper, we unlock the potential of text prompts by repositioning them as a direct modulator of video embeddings for controllable V2A.

\begin{figure*}[t!]
  \centering
  \includegraphics[width=\textwidth]{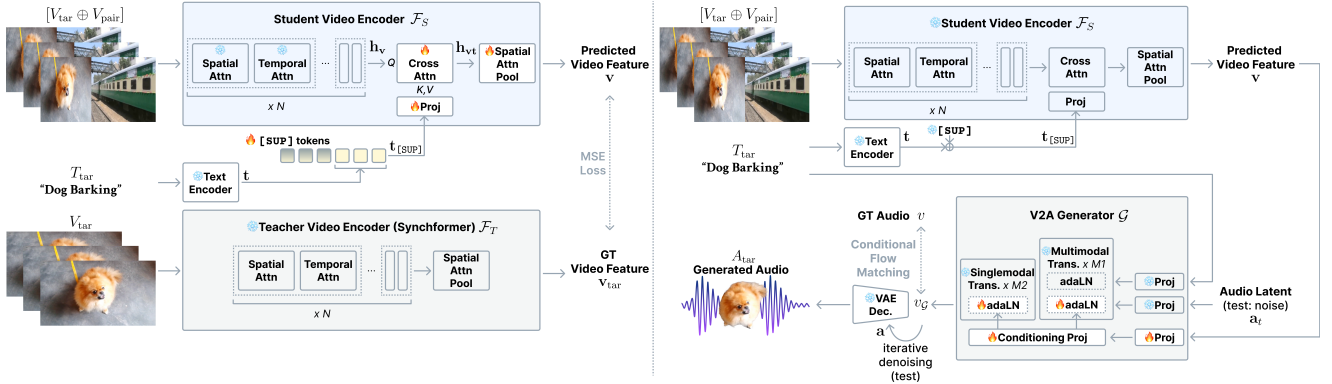}
  \vspace{-2em}
  \caption{The overall training pipeline of \ours. We learn a text-conditioned video encoder with a teacher-student distillation manner (left; first stage), and train an audio generator that conditions on text and isolated visual cues for the sound source (right; second stage).
  $\learn{\text{Learnable}}$ layers are marked with \fire, while $\frozen{\text{frozen}}$ layers are marked with \snowflake.}
   \label{fig:train}
\end{figure*}

\textbf{Selective sound generation} has only recently begun to emerge for professional multimedia production, where models synthesize audio exclusively for target sound sources.
Hayakawa~\etal~\cite{negativeaudioguidance} proposed an iterative, track-wise approach with sequential generation, where sounds produced in previous steps are excluded from the current one.
This is accomplished using negative audio guidance that steers the flow-matching process to avoid regenerating audio from prior stages.
While the motivation is related to ours, their method heavily relies on the limited separation capability of the pretrained V2A model, especially at the first generation stage.
Otherwise, some works~\cite{saganet,hearyourclick} have utilized visual region-level cues, such as segmentation masks produced by pretrained models (\eg, SAM2~\cite{sam2}), for object-focused sound generation. 
However, those approaches have notable limitations in that they necessitate the integration of computationally expensive pretrained segmentation models~\cite{sam2, deva}, which have often struggled with occluded objects or incorrectly identifying non-object sounding sources (\eg, rain drop, wind blowing).
To overcome those limitations, our \ours is the first to introduce a text prompt for describing the target sound source within the input video for robust selective sound generation.

\section{Method}
\subsection{Motivation and problem statement}
While existing works~\cite{mmaudio,diff-foley,rewas} have achieved promising results in generating holistic sounds aligned with the input video, they often suffer from low fidelity to the text prompts.
Specifically, the model occasionally produces undesired outputs, \ie, non-target objects' audio that appear in the video but are not specified by the text prompt.
This limitation arises mainly because most approaches directly feed video features extracted from a frozen visual encoder—typically pretrained for general recognition tasks—into the generation pipeline. 
Such visual features tend to be noisy and entangled, containing both irrelevant visual cues alongside sound-related semantics. 
As a result, selectively generating only the intended sound remains challenging.
This hinders users from creating harmonious audio in real-world scenarios~\cite{oh2023demand,challengetta,negativeaudioguidance}. 
For example, a professional audio creator often needs to synthesize a soundtrack with various elements such as speech, music, and sound effects under separate
controllable conditions.
At this juncture, we argue that the text-conditioned video feature grounding could make a huge room for the controllability of V2A.

Given a video $V$ paired with an audio $\bar{A}=\sum_{i}{A_i}$ which is a mixture of multiple sound sources, and a text prompts $\{T_i\}$ that describes a $i$-th specific sound source, \ours aims to generate audio exclusively that corresponds to the text:
\begin{equation}
    A_i = \mathcal{G}(\mathcal{F}(V, \bt_i), \bt_i)
\end{equation}
where $\mathcal{F}$ is a visual encoder and $\mathcal{G}$ is a generative model, and a text feature $\bt_i={\mathcal{E}}(T_i)$ is obtained from a text encoder ${\mathcal{E}}$.
Text prompts role as explicit selectors for video features to offer two main advantages over visual prompts.
First, they clearly deliver the target sound source, while visual sound sources often fail to be segmented due to visual occlusions or camera movements. 
Second, text prompts offer flexible controllability, allowing users to modify the generated sound through simple language edits rather than complex visual manipulations.
Such editability supports intuitive control, facilitating practical use in post-production workflows.
Note that we employ a parameter-efficient tuning strategy, while most parameters are initialized from prior works and frozen.
In what follows, \learn{\text{learnable}} parameters appear in red, and \frozen{\text{frozen}} parameters appear in blue.


\subsection{Text-guided visual feature generation}
\paragraph{Cross-attention block}
\ours modulates visual features to encode sound-source relevant information that the text prompt describes.
Most V2A models~\cite{specvqgan,condfoleygen,foleycrafter,v-aura,mmaudio,multifoley,vaflow} rely on a pretrained vision encoder and, optionally, a text encoder to extract conditioning features.
The vision encoders are generally frozen during the training process, serving as visual feature extractors for audio generation.
The extracted visual features encompass the global scene context, yet they inherently carry noisy and excessive irrelevant information.
Thereby, it impedes the generation of user-intended sound.

Our goal is to produce text-aligned video features by efficiently finetuning the video encoder $\mathcal{F}$.
The base encoder is Synchformer~\cite{synchformer}, which is commonly used in recent V2A models~\cite{mmaudio,rewas,v-aura,hunyuanvideo}.
We introduce two key techniques: (1) A text-guided cross-attention block is inserted after the frozen spatiotemporal attention blocks to modify intermediate visual features relevant to the text guidance, with only small extra parameters.
Text features, obtained by pretrained text encoder $\frozen{\mathcal{E}}$ (\eg, FLAN-T5-Base~\cite{flant5}), are employed as keys and values in the attention process.
Formally, given a hidden video embedding after spatial and temporal attention blocks, $\mathbf{h_v}$, and a text embedding $\bar{\bt}=\learn{\texttt{Proj}}(\frozen{\mathcal{E}}(T))$ from the text encoder followed by a projection layer, the cross attention is performed: 
\begin{equation}
    \mathbf{h_{vt}}=\learn{\texttt{Cross-Attn}}(Q=\mathbf{h_v},K=\bar{\bt},V=\bar{\bt}).
    \label{eqn:crossattn}
\end{equation}
The predicted video feature $\bv$ is obtained with a learnable spatial attention pooling layer (\learn{\texttt{Spatial-Attn-Pool}}).

\begin{figure}[t!]
    \centering
    \begin{subfigure}[]{0.5\linewidth}
        \centering 
        \includegraphics[width=\linewidth]{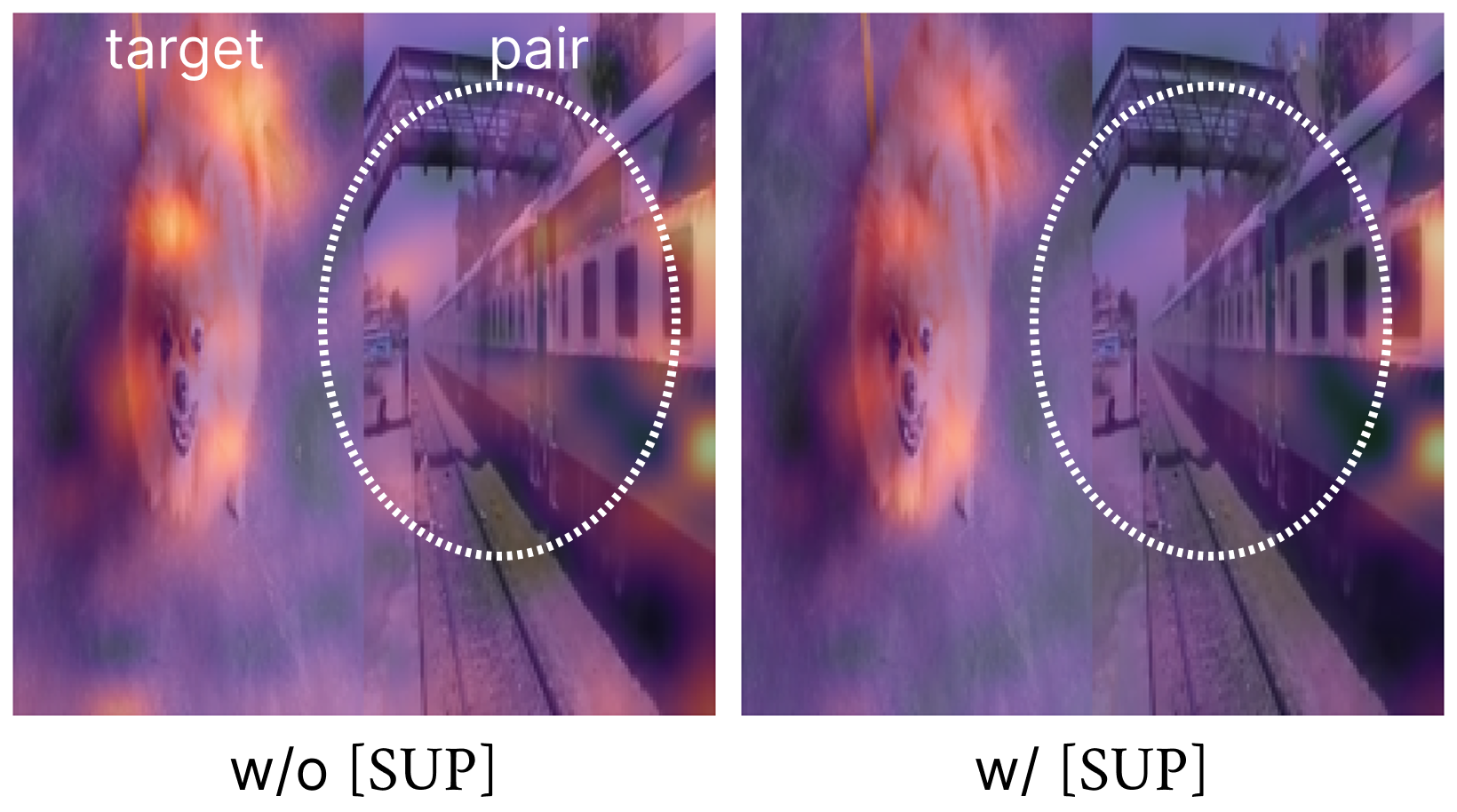}
        \caption{``dog barking"}
        \label{fig:attn_dog}
    \end{subfigure}\hfill
    \begin{subfigure}[]{0.5\linewidth}
        \centering
        \includegraphics[width=\linewidth]{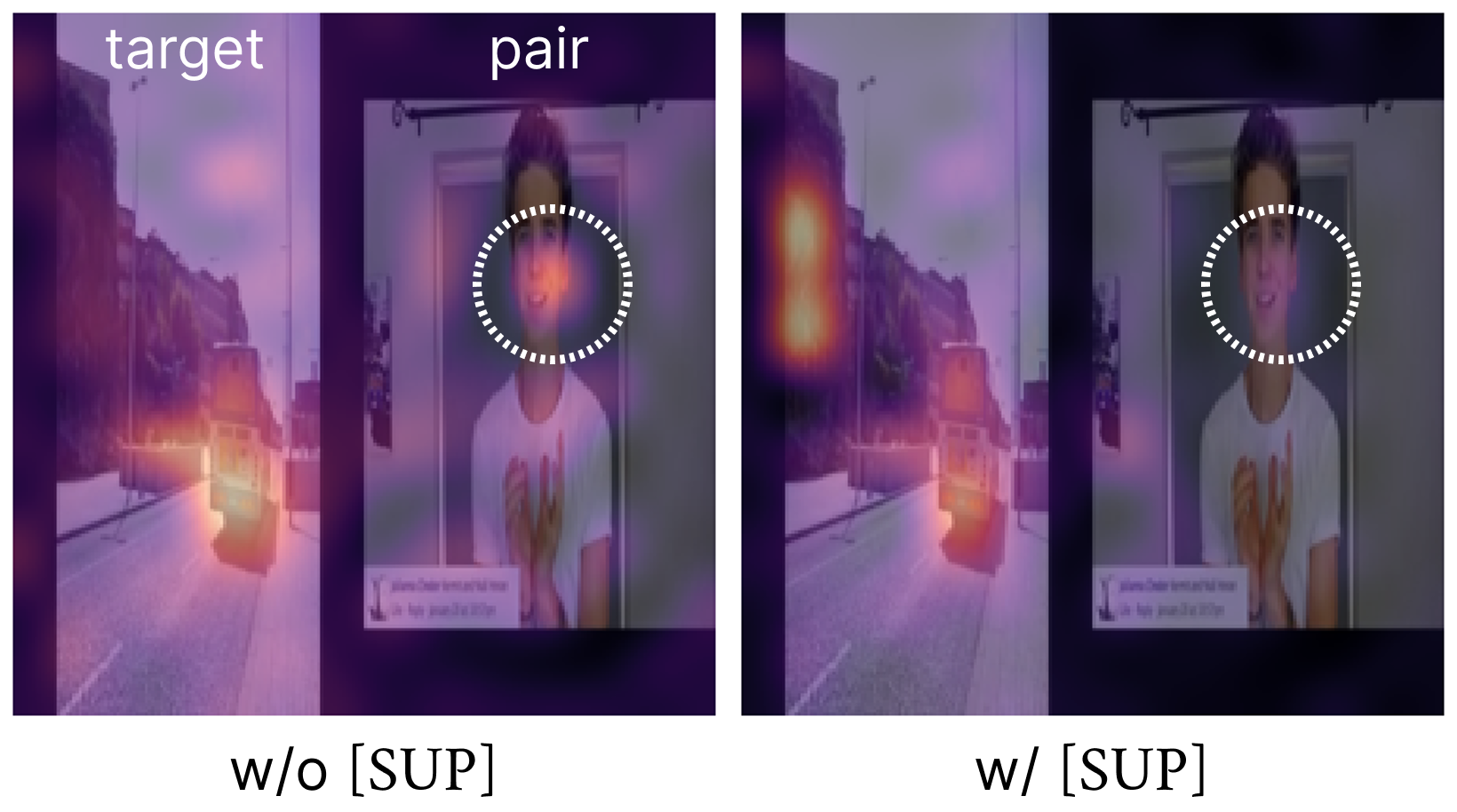}
        \caption{``driving buses"}
        \label{fig:attn_bus}
    \end{subfigure}\hfill
    \begin{subfigure}[]{0.5\linewidth}
        \centering
        \includegraphics[width=\linewidth]{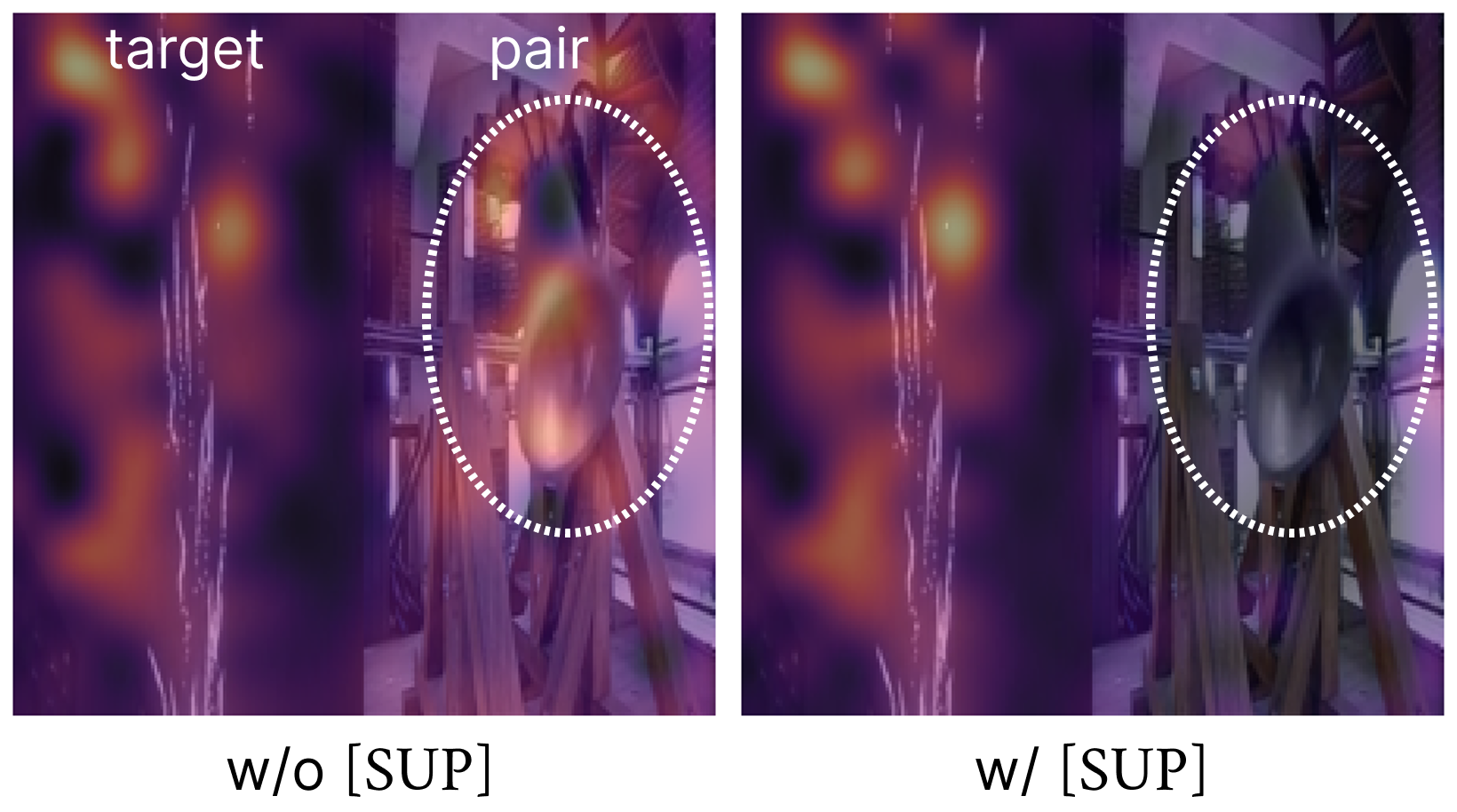}
        \caption{``firework banging"}
        \label{fig:attn_firework}
    \end{subfigure}\hfill
    \begin{subfigure}[]{0.5\linewidth}
        \centering
        \includegraphics[width=\linewidth]{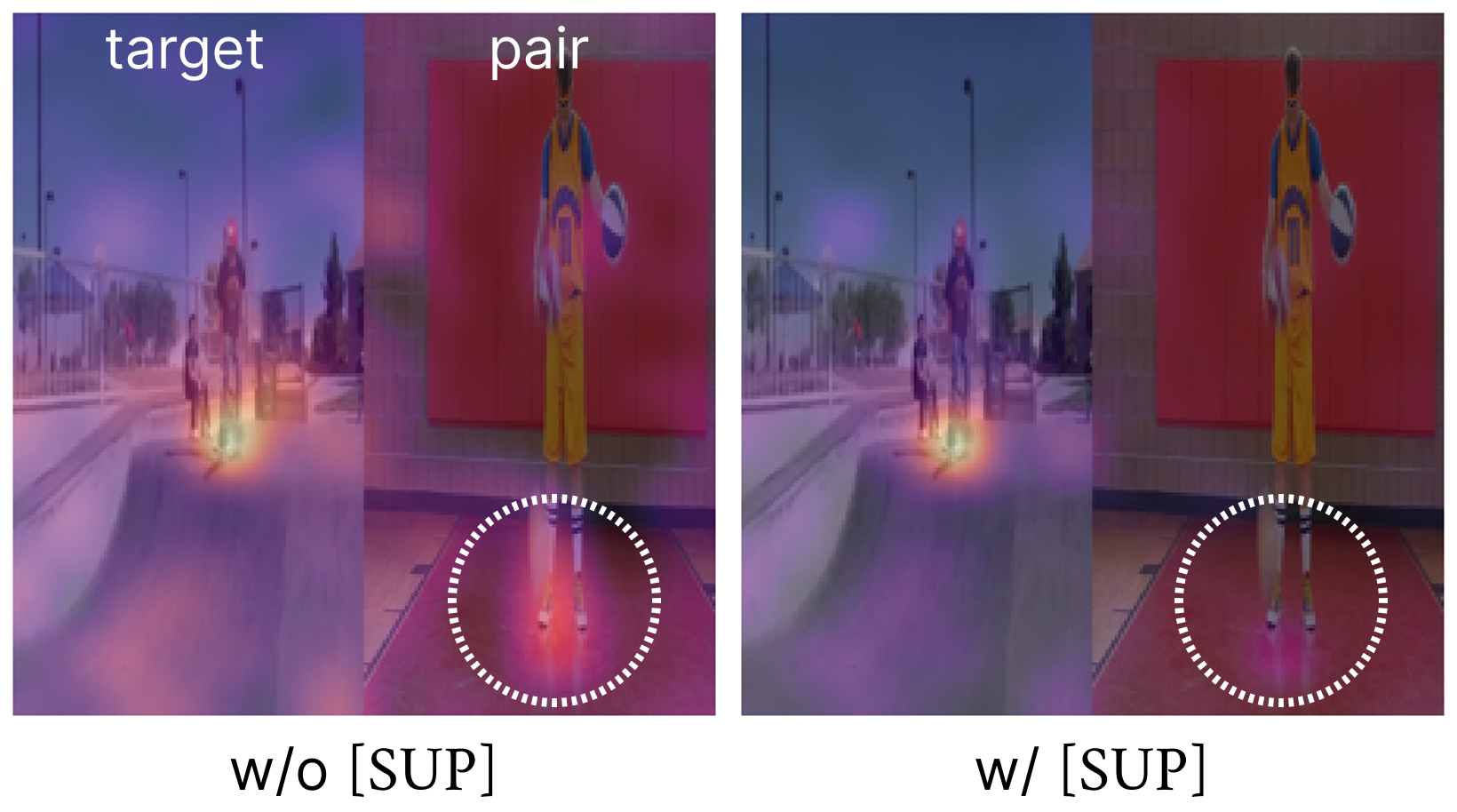}
        \caption{``skateboarding"}
        \label{fig:attn_skateboard}
    \end{subfigure}\hfill
    \vspace{-.5em}
    \caption{Attention visualization for \texttt{[eos]} token over auto-mixed frame in the last block without (left) / with (right) \reg tokens. Each subcaption denotes the corresponding target prompt.}
    \label{fig:attn_eos}
\end{figure}
\paragraph{Learnable supplementary tokens}
We expect hidden video embeddings to contain semantics that are exclusively aligned with text. 
However, a simple cross-attention mechanism yields a suboptimal result where the model still suffers from generating sounds corresponding to the motion dynamics of non-target instances.
For example, when a dog is barking beside a cat, it incorrectly produces a meowing sound, reflecting semantic confusion between co-occurring sound sources.
To mitigate this issue, we introduce a novel technique to attach learnable supplementary tokens \learn{\reg} preceding text embeddings. 
As shown in \cref{fig:attn_eos} (`w/o \reg'), such artifacts emerge as semantic patch tokens (particularly those associated with motion dynamics) become high-norm outliers in the attention space. 
However, as in prior approaches on vision transformers~\cite{vit_register,cavmae_sync}, adding extra tokens to the sequence of visual embeddings increases the computational cost across all encoder blocks.
Furthermore, we should design to learn intent-focused visual representations that emphasize regions and cues relevant to the user’s specified sound source.
To this end, \learn{\reg} are simply prepended to text features as follows:
\begin{equation}
    \bt_{\reg}=[\learn{\reg} \oplus \bt],
\end{equation} 
where $\oplus$ is a sequence-wise concatenation operation.
The cross-attention block uses $\bt_{\reg}$ to produce intent-modulated video feature $\mathbf{h}_{\bv\bt}$ as in \cref{eqn:crossattn}.
This design enables the model to suppress irrelevant or misleading visual activations while strengthening attention toward regions that correspond to the user-intended sound source, as demonstrated in \cref{fig:attn_eos} (`w/ \reg').
As a result, the learned video representation leads to improved selectivity and audiovisual coherence.
The detailed input data configuration of the video encoder is in Appendix \ref{sec:sync_feat}.

\subsection{Selective sound generation}
Our generator, $\mathcal{G}$, adopts a multimodal diffusion transformer (MM-DiT) architecture~\cite{sd3,flux}:
\begin{equation}
    \ba = \mathcal{G}(\bv, \bt_\tar, \ba_\tar)
\end{equation}
where $\ba_\tar$ is the audio latent extracted from a pretrained variational autoencoder (VAE)~\cite{vae}.
It closely follows the pipeline of MMAudio~\cite{mmaudio}, with the only modification being the exclusion of CLIP~\cite{clip} features from the conditioning inputs.
It is worth noting that the generator already possesses sufficient capacity to synthesize audio conditioned on multimodal inputs.
Therefore, our contribution does not lie in designing a specialized generator architecture, but rather in enabling selective sound generation through improved conditioning representations.
Specifically, the initially sampled noise is transformed to target audio latent $\hat{\ba}$ via a flow matching process~\cite{cfm,cfm_ot}, jointly contextualizing video $\bv$ and text semantics $\bt$ in MM-DiT.
MM-DiT consists of a stack of multimodal and single-modal transformer blocks. 
Given hidden audio features $\mathbf{h}_\ba$ and text features $\mathbf{h}_\bt$, the multimodal blocks compute $\frozen{\texttt{Self-Attn}}(Q,K,V=[\mathbf{h}_\bt,\mathbf{h}_\ba])$, while the single-modal blocks compute $\frozen{\texttt{Self-Attn}}(Q,K,V=\mathbf{h}_\ba)$.
The adaptive LayerNorm ($\learn{\texttt{adaLN}}$) layers~\cite{peebles2023scalable} condition the block-wise hidden state $\mathbf{h}\in\mathbb{R}^{L\times d}$ on the linear-projected video feature $\bar\bv=\learn{W_\bv}\bv$. Formally, this operation is defined as: 
\begin{equation}
    \learn{\texttt{adaLN}}(\mathbf{h},\bar\bv)=\mathbf{1} \learn{W_\gamma}(\bar\bv)\cdot\texttt{LN}(\mathbf{h})+\mathbf{1}\learn{W_\beta} (\bar\bv)
\end{equation} where $\learn{W_\gamma}$ and $\learn{W_\beta}$ are the conditioning projection layers and $\mathbf{1}\in\mathbb{R}^{L\times1}$ is a matrix of ones for broadcasting. 

\subsection{Training}
\paragraph{Video-mixing} 
As input videos usually comprise multiple sound sources, without explicit annotations for each sound source, it is nontrivial to isolate visual features.
To address this, we introduce a self-supervised strategy, motivated by audiovisual separation works~\cite{ephrat2018looking,lee2021looking}, but reformulated for selective V2A generation.
Concretely, two videos are randomly selected, and horizontally concatenated with a random ratio to make the desired \{mixed-video, audio, text\} pairs.
One of the audio-text pairs is randomly chosen to serve as the target. 
Formally, an input video $V$ in a mini-batch consists of randomly selected two videos $\{V_\tar, V_\pair\} \in \mathbb{R}^{H\times W}$:
\begin{equation}
    V =[V_\tar \in \mathbb{R}^{H\times \lambda W} \oplus V_\pair \in \mathbb{R}^{H\times (1-\lambda) W}],
\end{equation}
where $\lambda \sim \text{Beta}(\alpha, \alpha)$ is a scaling factor for resizing the video sampled from a beta distribution, and $\oplus$ is a horizontal concatenation operation.
Here, $V_\tar$ serves as a target video to semantically attend, while the paired video $V_\pair$ becomes a distractor.
This scheme encourages robust cross-modal grounding by distinguishing the target visual region without explicit supervision.

\paragraph{Two-stage training} 
Learning \textit{conditional feature extraction from mixed sources} and \textit{multiple conditioned audio generation} simultaneously is inherently complex, as both modules depend on each other's evolving representations~\cite{he2020momentum}.
To ensure efficient and stable optimization, the joint training is organized into a two-stage learning scheme, allowing each module to converge toward a consistent representation before mutual conditioning.
In the first stage, the video encoder learns to understand text prompts from output features of the teacher model~\cite{zhang2019your,noisy_student}.
In the left part of \cref{fig:train}, while the teacher model $\mathcal{F}_{T}$ (\ie, pretrained Synchformer~\cite{synchformer}) takes a single source video $V_\tar$ to generate a pseudo feature $\bv_\tar$, the student model $\mathcal{F}_S$ produces a text-guided visual feature from the mixed source.
Formally, the video feature $\bv$ of student encoder $\mathcal{F}_S$ is learned to minimize the L2-norm regression loss:
\begin{equation}
     ||\mathcal{F}_{S}([V_\tar \oplus V_\pair], \bt_\tar)-\mathcal{F}_{T}(V_\tar)||^2 
    \label{eq:teacher_student}
\end{equation}
where $\bt_\tar$ is the text embeddings corresponding to the target video $V_\tar$.
This stage updates the student network's parameter for cross-attention and spatial attention pooling layers exclusively.
While the teacher model can only take visual inputs, the student extracts a specific representation from text guidance.
In other words, our approach uses video features as teaching signals, allowing the model to learn how multimodal interactions can selectively emphasize informative cues while suppressing irrelevant sound sources as noise in the visual representation.


Next, we train a generator $\mathcal{G}$ while keeping the video encoder frozen in the second stage. 
We start from the MM-DiT in MMAudio~\cite{mmaudio} as the baseline of the generator.
Rather than finetuning the whole parameters, we focus on specific modules that handle video features explicitly, as illustrated in the right part of \cref{fig:train}. 
Specifically, we finetune two sub-modules only: (1) the initial projection layer of the video feature branch (\ie, $W_\bv$) and (2) the adaptive LayerNorm (adaLN) module of the audio latent branch of multimodal and single-modal transformer blocks (\ie, $W_\gamma, W_\beta$). 
The model is trained with conditional flow-matching (CFM)~\cite{cfm,cfm_ot}. 
Given a noise distribution $q(\ba_0)\sim\mathcal{N}(\mathbf{0},I)$, training data distribution $q(\ba_1=\ba_\tar,\bc)$ with input condition features $\bc=(\hat\bv_\tar=\mathcal{F}_S(V,\bt_\tar),\bt_\tar)$, and timestep $t\in[0,1]$, the CFM objective is formulated as:
\begin{equation}
     \mathop{\mathbb{E}}_{t,q(x_0),q(\ba_1,\bc)}||v(t,\bc,\ba_t; \mathcal{G})-v(\ba_t|\ba_0,\ba_1)||^2,
     \label{eq:cfm}
\end{equation}
where $\ba_t=t\ba_1+(1-t)\ba_0$ is the flow that generates a velocity $v$. 
Implementation details for optimizer settings are noted in Appendix \ref{sec:train_detail}.


\section{Experiments}

\begin{table*}[t!]
\setlength{\tabcolsep}{1em}
\centering
\small
\begin{tabular}{lccccccc}
\toprule
 \multirow{2}[3]{*}{Model}&
  \multicolumn{3}{c}{\cellcolor{backGray}Audio Quality} &
  \multicolumn{3}{c}{\cellcolor{backBlue}Semantic Alignment} &
  {\cellcolor{backRed}Temporal Alignment} \\
 \cmidrule(l{1pt}r{1pt}){2-4}
 \cmidrule(l{1pt}r{1pt}){5-7}
 \cmidrule(l{1pt}r{1pt}){8-8}
 &
  FAD$\downarrow$ &
  KAD$\downarrow$ &
  IS$\uparrow$ &
  KL$\downarrow$ &
  CLAP$\uparrow$ &
  IB$\uparrow$ &
  DeSync$\downarrow$ \\ \midrule
\textcolor{gray}{\textit{Inter-class}} \\
ReWaS~\cite{rewas} & 70.4 & 4.937 & \hspace{0.5em}6.23 & 2.57 & 0.200 & 0.2454 & 1.364\\
VinTAGe~\cite{vintage} & \textbf{50.5} & 1.309 & 11.51 & \textbf{1.69} & 0.283 & 0.2850 & 1.292\\
MMAudio-S-16k~\cite{mmaudio} & 56.7 & \underline{0.874} & 11.54 & 2.07 & 0.270 & 
\underline{0.3135} & \underline{0.802} \\
VOS~\cite{deva}+MMAudio~\cite{mmaudio} & 60.0 & 0.878 & \underline{12.11} & 1.91 & \underline{0.291} & 0.3010 & 0.991 \\ 
\ours & \underline{51.7} & \textbf{0.676} & \textbf{13.07} & \underline{1.85} & \textbf{0.292} & \textbf{0.3251} & \textbf{0.721} \\ \midrule
\textcolor{gray}{\textit{Intra-class}} \\
ReWaS~\cite{rewas} & 57.4 & 3.148 & 6.29 & 1.97 & 0.220 & 0.2569 & 1.377\\
VinTAGe~\cite{vintage} & \textbf{37.0} & 0.690 & \underline{9.28} & \textbf{0.88} & 0.277 & 0.2892 & 1.304 \\
MMAudio-S-16k~\cite{mmaudio} & \underline{41.5} & \underline{0.654} & 9.00 & 1.09 & 0.276 & \underline{0.3248} & \underline{0.670} \\
VOS~\cite{deva}+MMAudio~\cite{mmaudio} & 43.4 & 0.656 & 8.91 & 1.11 & \textbf{0.287} & 0.3087 & 0.904 \\ 
\ours & \textbf{37.0} & \textbf{0.492} & \textbf{9.62} & \underline{1.04} & \underline{0.280} & \textbf{0.3262} & \textbf{0.639} \\ 
\bottomrule
\end{tabular}%
\vspace{-.5em}
\caption{Quantitative comparisons with state-of-the-art models on \ourbench. All methods used text prompts corresponding to the target videos. The \textbf{best} scores are shown in bold, and the \underline{second-best} scores are underlined.}
\label{tab:objective_eval}
\end{table*}

\subsection{Setup}
\label{ssec:exp_setup}
\paragraph{Training dataset} 
\ours is trained on VGGSound~\cite{vggsound}, which provides approximately 500 hours of video and 310 unique captions. We utilize these captions as the text prompts for our model. Following the experimental setup of our baseline model~\cite{mmaudio}, we partition the official training data, setting aside 2k samples for validation. The training set includes 179k videos, and the test set is 15k.
For both training and inference, all video clips are processed into 8-second segments.

\paragraph{Test benchmark}
Evaluating selective sound generation requires clean, source-separated audio with corresponding text descriptions.
However, existing in-the-wild datasets such as VGGSound~\cite{vggsound} and AudioSet~\cite{audioset} typically provide only a single mixed track and video-level captions, often contaminated by recording noise or off-screen sounds~\cite{regnet,varietysound,vintage}. 
To address these limitations, we introduce \textbf{\ourbench}, an evaluation benchmark for selective V2A generation. 
We curate mono-source clips from UnAV-100~\cite{unav100} overlapping with VGGSound test set, and filter them automatically and manually with three strict criteria: (1) a single source sounding with minimal background or off-screen noise, (2) the sounding object is clearly visible, (3) the text annotation precisely matches the auditory event.
Finally, we obtain a final set of 67 clean, single-source videos spanning 39 unique events (\eg, `dog barking', 'train wheels squealing') across 8 categories: \textit{human, music, vehicle, tool, animal, nature, sport, other}.
To construct test samples, we concatenate pairs of these videos side-by-side, each occupying half the frame width. 
The horizontally combined video $[V_{1} \oplus V_{2}]$ serves as the model input, while the audio $A_{\bv_1}$ and text $T_{\bv_1}$ from one of the source videos are used as the target. 
From the 67 curated videos, we generate 1,071 mixed pairs in total, 560 inter-class (videos from different categories) and 511 intra-class (videos from the same category), for quantitative evaluation.
Appendix \ref{sec:benchmark_apdx} provides the full list of audio event categories and detailed benchmark statistics.
We also include results on the original VGGSound test set for completeness; these are given in Appendix~\ref{sec:apdx_vggsoundtest}, as this setting is not central to our evaluation.

\paragraph{Baselines}
We establish four SoTA baselines for comparison. 
ReWaS~\cite{rewas}, VinTAGe~\cite{vintage}, and MMAudio~\cite{mmaudio} are text-conditioned V2A models where text semantically aids video.
To implement concurrent segmentation-based approaches~\cite{hearyourclick,saganet}, we leverage the pretrained video object segmentation (VOS) model~\cite{deva} to build a `VOS+MMAudio' system, in which we pass the text and video to obtain video-level object masks. 
The resulting masked video is then used as the conditional input to the MMAudio. 

\begin{figure*}[t!]
    \centering
    \begin{subfigure}[]{0.5\textwidth}
        \centering 
        \includegraphics[width=\linewidth]{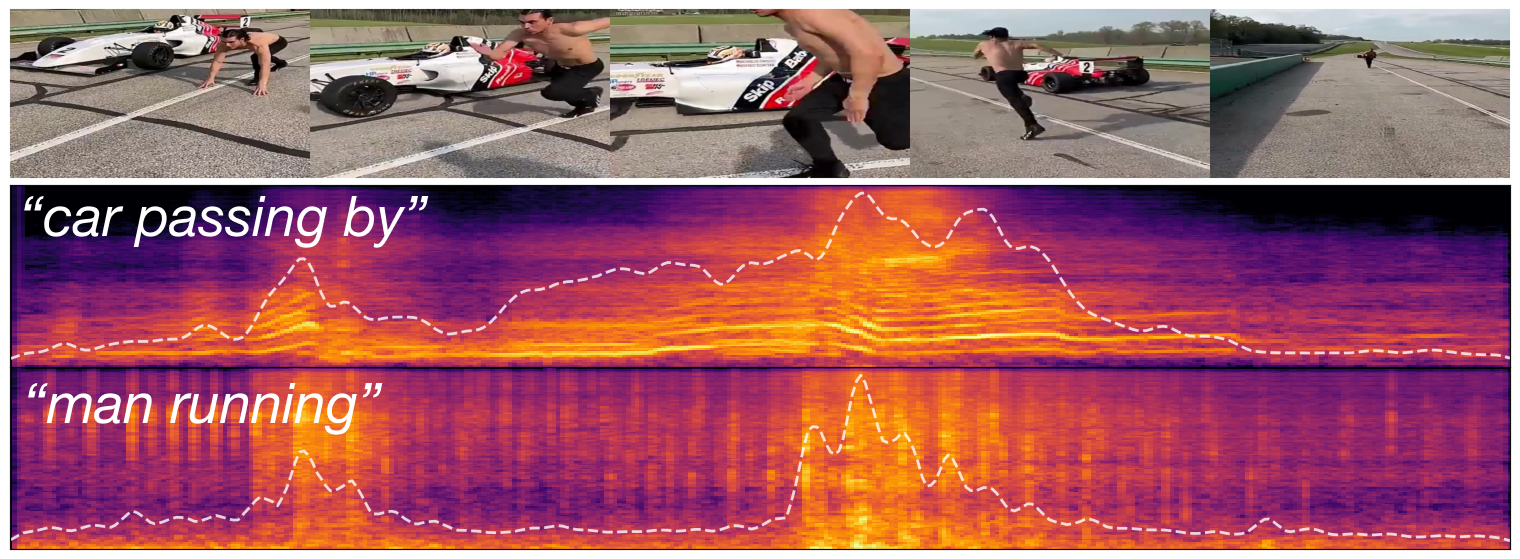}
        \caption{Spatially co-occurring events.}
        \label{fig:eg_spatial}
    \end{subfigure}\hfill
    \begin{subfigure}[]{0.5\textwidth}
        \centering
        \includegraphics[width=\linewidth]{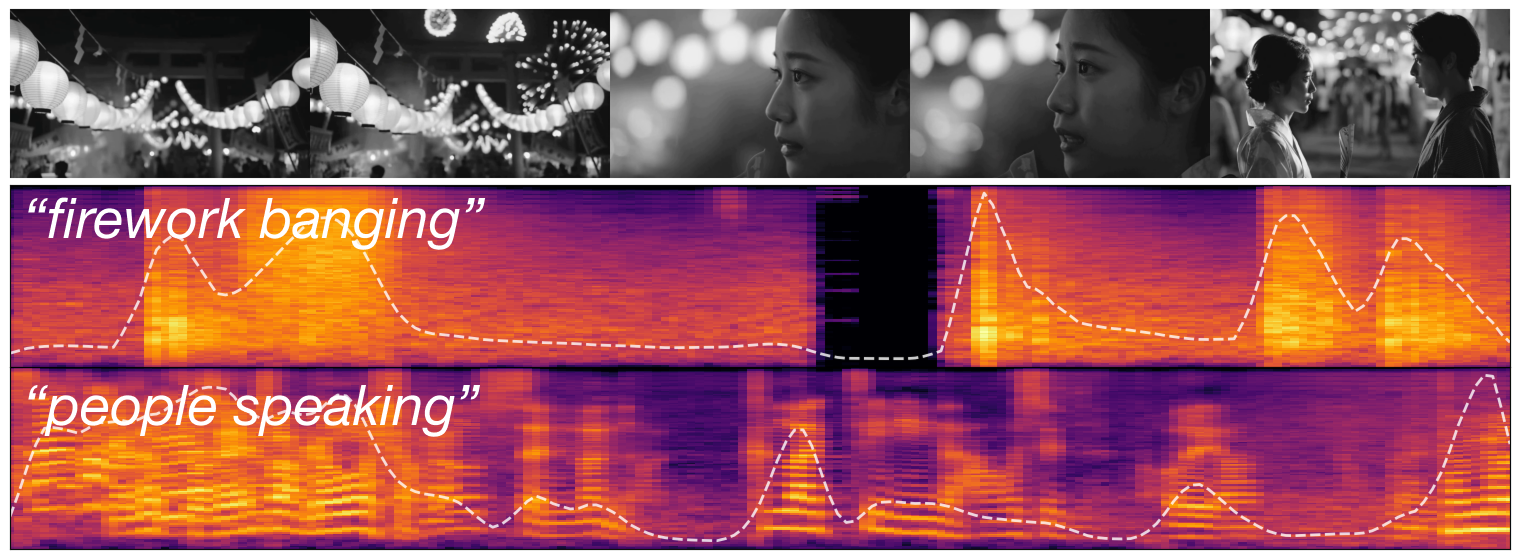}
        \caption{Multiple events intertwined temporally.}
        \label{fig:eg_temporal}
    \end{subfigure}\hfill
    \vspace{-.5em}
    \caption{Examples of selective generation with real-world  videos. The white dotted curve is the root-mean-squared audio amplitude.}
    \label{fig:qual_eg}
\end{figure*}

\paragraph{Metrics}
Three main criteria matter for evaluating selective audio generation: audio quality, semantic alignment, and temporal alignment with the target.
\begin{itemize}
    \item \underline{Audio quality} is assessed by Fréchet audio distance (FAD)~\cite{fad}, kernel audio distance (KAD)~\cite{kad}, and inception score (IS)~\cite{inception_score} with PANNs\cite{panns,correlation_fad}.
    \item \underline{Semantic alignment} is assessed to evaluate prompt fidelity.
    While CLAP score (CLAP)~\cite{clap_laion} is used to measure how closely the generated audio aligns with the intended text, imagebind score (IB)~\cite{imagebind} measures the alignment between audio and target video.
    In addition, Kullback-Leibler divergence (KL) with PANNs distribution is employed to evaluate semantic alignment between the generated and groundtruth audio tracks.
    \item \underline{Temporal alignment} is assessed by DeSync~\cite{mmaudio}, the average synchronized error (\ie, predicted offset in seconds) between the audio and video.
    As temporal alignment is crucial for perceptual coherence in V2A, this metric serves as the primary reference in our ablation study.
\end{itemize}

\subsection{Implementation details}
During training of \ours, mixing inputs are given within each minibatch with a probability of 0.75, while clipping the mixing ratio $\lambda$ of the target video to be greater than 0.2. 
A total of 5 learnable \reg tokens are prepended to every input text prompt; this number was determined by \cref{tab:ablation_token}.
We initialize the video encoder $\mathcal{F}_S$ with pretrained Synchformer~\cite{synchformer} and the generator $\mathcal{G}$ with MMAudio-small-16kHz weights.
Note that we train 19M parameters in $\mathcal{F}_S$ and 22M in $\mathcal{G}$, corresponding to 14\% of each model's total parameters, respectively.
Following the original setup for classifier-free guidance (CFG) in MMAudio~\cite{mmaudio}, we randomly substitute the video and text features with learned null video and text embedding ($\varnothing_\mathbf{v}$ and $\varnothing_\mathbf{t}$) with a probability of 0.1. 
In addition, we drop the text feature with an additional probability of 0.5 to enhance the visual fidelity.
Inference on the flow matching model is performed using the Euler solver with 25 linear sampling steps. 
During inference, CFG is applied with a guidance strength of $\gamma=4.5$.

\subsection{Comparison with state-of-the-arts}


\paragraph{Quantitative analysis}
Table \ref{tab:objective_eval} summarizes the quantitative performance of V2A models on \ourbench. 
\ours outperforms baselines across all key aspects, including audio quality, semantic alignment, and temporal alignment. 
Notably, we achieve the best scores in both semantic and temporal audio-video alignment.
MMAudio~\cite{mmaudio}, which overlooks text modality, exhibits degraded CLAP scores than \ours, whereas neglecting video modality often results in temporally misaligned results with poor DeSync scores, as seen in ReWaS~\cite{rewas} and VinTAGe~\cite{vintage}. 
This highlights that training a text-conditioned video encoder in \ours is effective to achieve these dual goals. 
VOS baseline shows competitive semantic alignment but performs poorly on temporal synchronization.
It is primarily due to the inherent limitations of VOS methods, which struggle to accurately localize fast-moving or motion-blurred objects, and vague or complex boundaries (\eg, rain drop). 
Regarding the two subsets of \ourbench, models generally achieve better objective scores in the intra-class subset. 
This happens because the paired non-target video is semantically similar to the target, leading objective metrics to overestimate model performance.
Therefore, human perceptual evaluation becomes essential.

\paragraph{Qualitative analysis}
\cref{fig:qual_eg} demonstrates that \ours selectively synthesizes target sounds even in complex real-world auditory scenes involving multiple simultaneous or temporally overlapping events.
As shown in \cref{fig:eg_spatial}, \ours successfully generates distinct sounds for spatially co-occurring sound events such as `car passing by' and `man running', demonstrating spatial disentanglement within a shared visual context. 
In \cref{fig:eg_temporal}, \ours succeeds in temporally disentangling intertwined events, producing natural and temporally aligned sounds for `firework banging' and `people speaking'.
These examples highlight our model’s robustness to yield realistic and context-aware audio by capturing the user’s intended focus.
More qualitative examples are provided in  \cref{fig:qual_bench} in Appendix.

\begin{figure}[t!]
    \centering
    \includegraphics[width=\columnwidth]{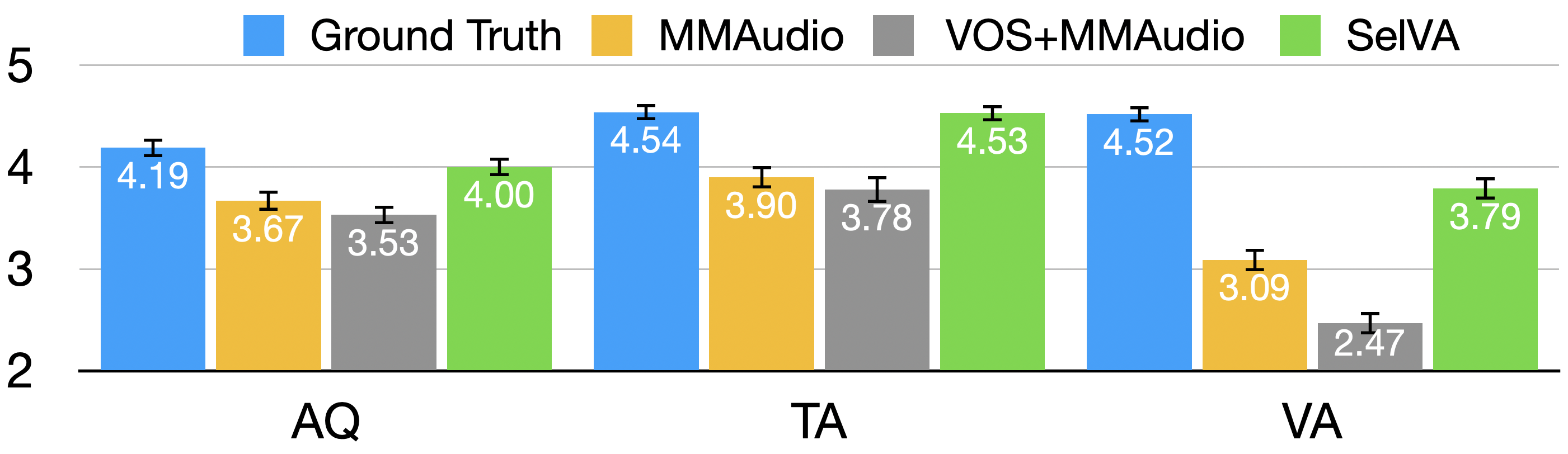}
    \vspace{-2em}
    \caption{Human study results on \ourbench. The GT results (\ie, real sound) show oracle performance. \ours outperforms state-of-the-art methods, including MMAudio and VOS baselines.}
    \label{fig:subjective_eval}
\end{figure}

\paragraph{Human study}\label{sec:human_study}
Human listening test assesses the perceptual performance of the models, to complement our automatic metrics. 
A total of 26 participants rated three criteria scores: \textit{overall audio quality} (AQ), \textit{text-audio alignment} (TA) for semantic relevance, and \textit{audiovisual temporal synchronization} (VA) using a 5-point Likert scale.
The evaluation set consists of 16 unique videos: one from each of the 8 sound categories, selected from both the inter-class and intra-class \ourbench benchmarks. 
Each video was presented with the corresponding audio by 4 different sources: ground-truth (GT), `MMAudio-S-16k', `VOS+MMAudio', and \ours. 
\cref{fig:subjective_eval} reports the mean opinion score (MOS), along with the corresponding 95\% confidence interval.
The subjective results show strong alignment with the objective evaluation.
\ours outperforms both MMAudio and the VOS baselines across all criteria.
In terms of audio quality, \ours achieves a comparable performance to GT, whereas the other baselines show noticeably lower scores.
For video-audio alignment, \ours also achieves the highest score among the comparable models, with the GT obviously achieving the best score.
Notably, VOS baseline scored 3.78 (vs. 4.53 in \ours) in text-audio alignment, even though its CLAP score in \cref{tab:objective_eval} is comparable to ours.
This highlights a discrepancy between the objective metric and human perception. 
It indicates that human listeners are more sensitive to off-screen noises that are loosely aligned with the text prompt (see Appendix \ref{sec:apdx_clap}).

\begin{table*}[t!]
\centering
\small
\begin{tabular}{lccccccc}
\toprule
 \multirow{2}[3]{*}{Model}&
  \multicolumn{3}{c}{\cellcolor{backGray}Audio Quality} &
  \multicolumn{3}{c}{\cellcolor{backBlue}Semantic Alignment} &
  {\cellcolor{backRed}Temporal Alignment} \\
 \cmidrule(l{1pt}r{1pt}){2-4}
 \cmidrule(l{1pt}r{1pt}){5-7}
 \cmidrule(l{1pt}r{1pt}){8-8}
 &
  FAD$\downarrow$ &
  KAD$\downarrow$ &
  IS$\uparrow$ &
  KL$\downarrow$ &
  CLAP$\uparrow$ &
  IB$\uparrow$ &
  DeSync$\downarrow$ \\ \midrule
\textcolor{gray}{\textit{Inter-class}} \\
\ours & 51.7 & 0.676 & 13.07 & 1.85 & 0.292 & 0.3251 & \textbf{0.721} \\
\hspace{2em} $-$ Video Enc. FT & 53.8 & \underline{0.638} & \underline{13.35} & \textbf{1.75} & \textbf{0.300} & \underline{0.3303} & 0.868 \\
\hspace{2em} $-$ V2A Gen. FT & 56.6 & 0.721 & 12.94 & 1.89 & 0.293 & \textbf{0.3309} & \underline{0.736} \\
\hspace{2em} $-$ \reg tokens & \underline{51.4} & \textbf{0.637} & 12.95 & \underline{1.79} & 0.289 & 0.3272 & 0.756 \\ 
\hspace{2em} $-$ two-stage training & \textbf{51.3} & 0.707 & \textbf{13.78} & 1.81 & \underline{0.299} & 0.3138 & 0.823 \\
\midrule
\textcolor{gray}{\textit{Intra-class}} \\
\ours & 37.0 & 0.492 & 9.62 & 1.04 & 0.280 & 0.3262 & \textbf{0.639} \\ 
\hspace{2em} $-$ Video Enc. FT & 38.2 & \textbf{0.423} & \underline{10.15} & \textbf{1.01} & \textbf{0.291} & \underline{0.3294} & 0.734 \\
\hspace{2em} $-$ V2A Gen. FT & 39.4 & 0.553 & 9.35 & 1.06 & 0.281 & \textbf{0.3300} & \underline{0.651} \\
\hspace{2em} $-$ \reg tokens & \textbf{36.3} & 0.485 & 9.74 & \textbf{1.01} & 0.281 & 0.3277 & 0.676 \\
\hspace{2em} $-$ two-stage training & \underline{36.8} & \underline{0.456} & \textbf{10.18} & \underline{1.00} & \underline{0.283} & 0.3229 & 0.777  \\ 
\bottomrule
\end{tabular}%
\vspace{-.5em}
\caption{Ablation on model design variants: without video encoder training, generator training, \reg tokens, and two-stage training.}
\label{tab:ablation}
\end{table*}

\subsection{Ablation studies}

\paragraph{Impact of each training component} \cref{tab:ablation} summarizes our ablation studies, which demonstrate the impact of removing each training component: (1) video encoder $\mathcal{F}_S$ finetuning (first stage), (2) V2A generator $\mathcal{G}$ finetuning (second stage), (3) prepended \reg tokens (used during the first stage), and (4) two-stage training.
Finetuning only the V2A generator (while keeping the video encoder frozen) yields marginal gains in audio quality and semantic alignment, but causes a notable degradation in audiovisual temporal synchrony. 
We observe that the generator tends to develop an undesired shortcut behavior, producing sounds that align text semantics but drift from the actual video events.
Conversely, excluding the V2A generator finetuning significantly reduces overall audio quality. 
The results indicate that finetuning the generator with auto-mixing samples is necessary to obtain optimal performance.
The results reveal that removing \reg tokens especially deteriorates the temporal alignment score. 
This supports our hypothesis that \reg tokens facilitate selective generation by refining text-irrelevant spatial attention, while making a negligible sacrifice in audio quality and semantic alignment.
Finally, joint training (\ie, optimizing \cref{eq:teacher_student} and \cref{eq:cfm} simultaneously) shows notable drops in both semantic and temporal audiovisual alignment scores, indicating that the model fails to maintain coherent cross-modal correspondence.
For instance, in the intra-class benchmark, IB (0.3229 vs. 0.3248) and DeSync (0.777 vs. 0.670) scores are even worse than the frozen MMAudio baseline.
In particular, joint training often substitutes non-target sound events with text-aligned sounds, thereby deteriorating temporal synchronization.

\begin{figure}[t!]
    \centering
    \begin{subfigure}[]{0.5\linewidth}
        \centering 
        \includegraphics[width=\linewidth]{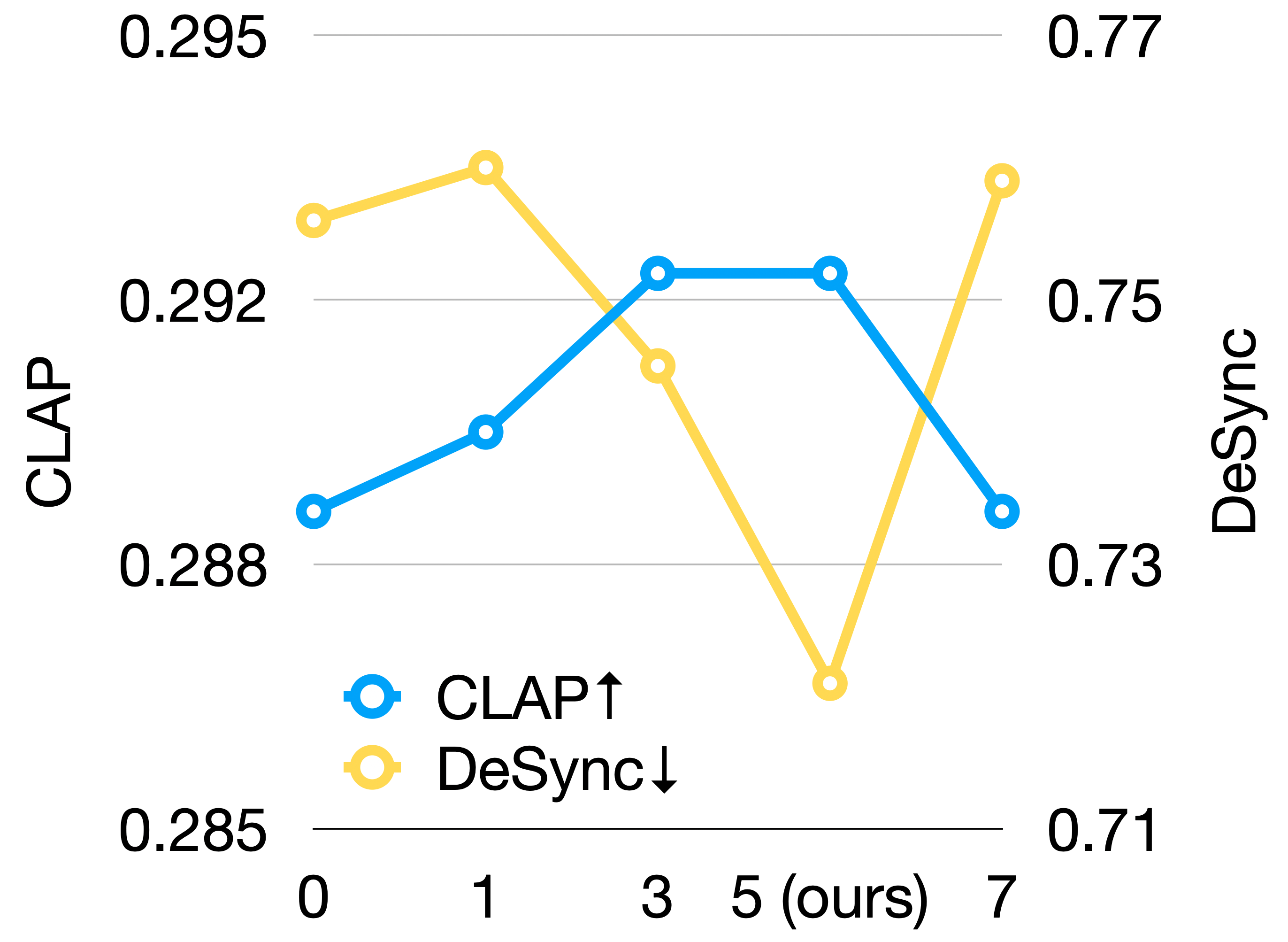}
        \caption{{Inter-class}}
        \label{fig:token_inter}
    \end{subfigure}\hfill
    \begin{subfigure}[]{0.5\linewidth}
        \centering
        \includegraphics[width=\linewidth]{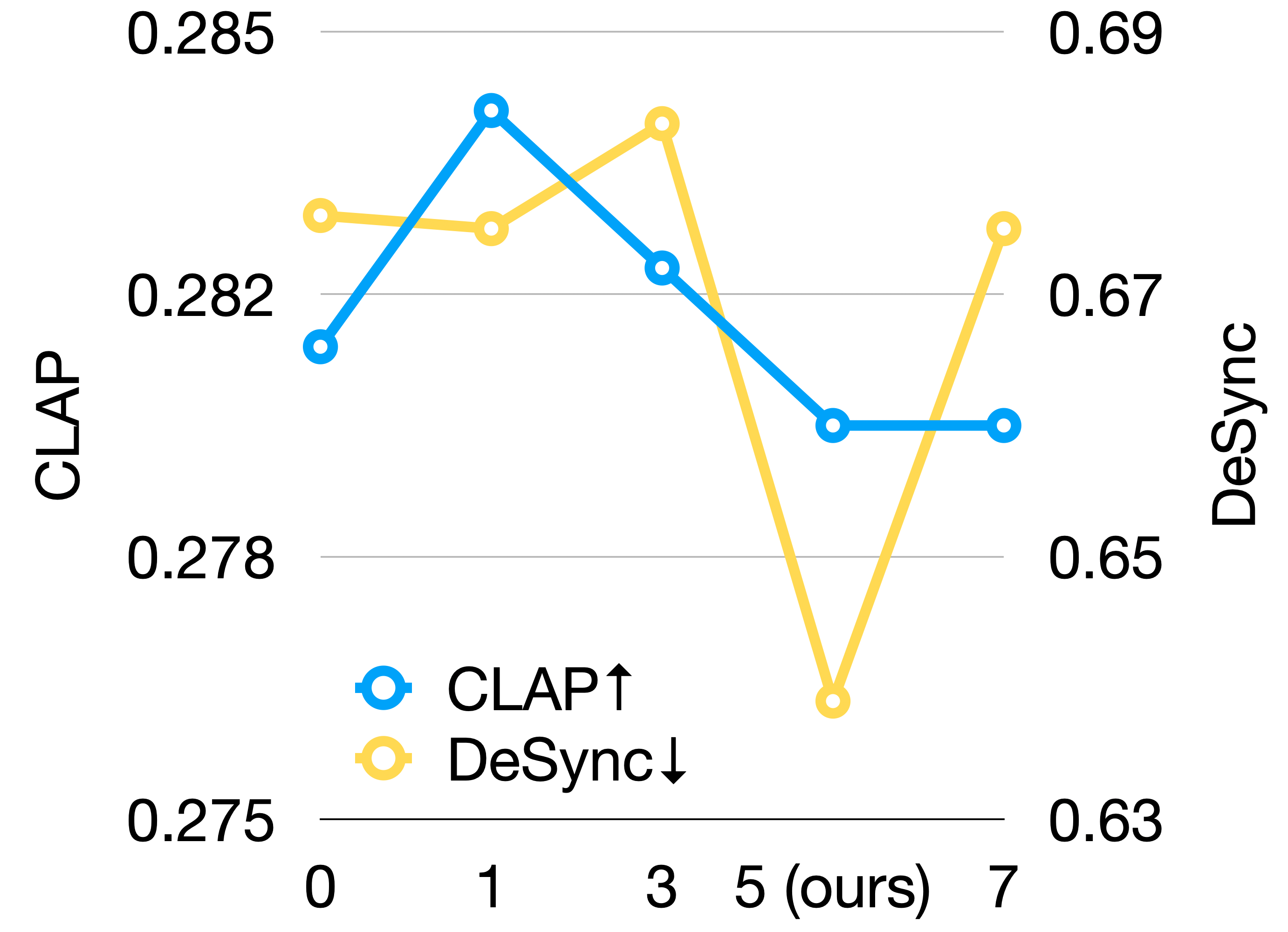}
        \caption{{Intra-class}}
        \label{fig:token_intra}
    \end{subfigure}\hfill
    \vspace{-.7em}
    \caption{Ablation on the number of \reg tokens, determined by balancing semantic and temporal alignment performance.}
    \label{fig:token_ablation}
\end{figure}
\paragraph{The number of \reg tokens} 
Achieving the dual goal of semantic text-audio alignment and temporal video-audio alignment is crucial in selective sound generation. 
We therefore observe the change in both the CLAP and DeSync scores for different numbers of \reg tokens. 
As shown in \cref{fig:token_ablation}, we identify a ``sweet spot" at 5 tokens, which achieves the best DeSync score while maintaining a comparable CLAP score. 
It is a consistent observation from prefix-tuning~\cite{prefix_tuning}, where performance typically degrades when too few tokens fail to convey sufficient conditioning information, or when too many tokens lead to redundancy and overfitting.
Full results, including other metrics, are shown in \cref{tab:ablation_token} in Appendix.

\section{Limitations and Future Work}
We identify three primary directions for future work.
First, the model performance is currently limited by the noisiness of the training data in VGGSound~\cite{vggsound}. 
Therefore, more rigorous data filtering or refining the auto-mixing process with cleaner source data could improve the performance.
Second, since text labels are typically simple noun-verb conjunctions and lack such descriptive detail, the model's complex text understanding capabilities could be enhanced. 
This includes fine-grained cross-modal distinction (\eg, separating `male singing' from `male burping') and improved attribute controllability (\eg, a dog barking `aggressively'). 
Finally, while our method significantly alleviates the sound substitution issue, residual cases remain when the video encoder fails to track a target movement change consistently. 
We leave a comprehensive full training of the model as future work.

\section{Conclusion}
We present \ours, text-conditioned V2A model tailored for audio production systems in the real world.
\ours efficiently modulates the video encoder to capture the user's textual intent, introducing a few learnable tokens and specialized training schemes.
Experimental results show that \ours delivers precise and controllable sound generation on our new benchmark, \ourbench, significantly outperforming existing methods.
These findings highlight \ours as a strong step toward practical, reliable, and fully controllable selective video-to-audio generation.

\section*{Acknowledgements} 
This work was supported by Institute of Information \& communications
Technology Planning \& Evaluation (IITP) grant funded by the Korea government (MSIT) (No.RS-2019-II190075, Artificial Intelligence Graduate School
Program (KAIST)), and the National Research Foundation of Korea (NRF) grant funded by the Korea government (MSIT) (No. RS-2023-00222383, RS-2025-16065706).
{
    \small
    \bibliographystyle{ieeenat_fullname}
    \bibliography{main}
}

\appendix
\setcounter{table}{0}
\setcounter{figure}{0}
\renewcommand\thetable{\thesection\arabic{table}}
\renewcommand\thefigure{\thesection\arabic{figure}}
\clearpage
\setcounter{page}{1}
\maketitlesupplementary


\section{More Implementation Details}
\begin{table*}[ht!]
\centering
\small
\begin{tabular}{lccccccc}
\toprule
 \multirow{2}[3]{*}{Model}&
  \multicolumn{3}{c}{\cellcolor{backGray}Audio Quality} &
  \multicolumn{3}{c}{\cellcolor{backBlue}Semantic Alignment} &
  {\cellcolor{backRed}Temporal Alignment} \\
 \cmidrule(l{1pt}r{1pt}){2-4}
 \cmidrule(l{1pt}r{1pt}){5-7}
 \cmidrule(l{1pt}r{1pt}){8-8}
 &
  FAD$\downarrow$ &
  KAD$\downarrow$ &
  IS$\uparrow$ &
  KL$\downarrow$ &
  CLAP$\uparrow$ &
  IB$\uparrow$ &
  DeSync$\downarrow$ \\ \midrule
MMAudio-S-16kHz~\cite{mmaudio} & \textbf{5.15} & \textbf{0.260} & 14.53 & \textbf{1.64} & 0.197 & \textbf{0.2927} & \textbf{0.486} \\
\ \ \ \ w/ null CLIP emb. & 7.85 & 0.338 & \textbf{18.95} & 1.75 & \textbf{0.235} & 0.2670 & 0.492 \\
\ \ \ \ w/ null Synchformer emb. & 7.51 & 0.563 & 14.27 & 2.00 & 0.196 & 0.2394 & 1.243 \\\bottomrule
\end{tabular}%
\vspace{-.7em}
\caption{Performance of MMAudio~\cite{mmaudio} on VGGSound test set with different input visual feature combinations.}
\label{tab:visemb}
\end{table*}
\subsection{Video encoder}
\paragraph{CLIP vs. Synchformer}\label{sec:vfeat_mmaudio}
We first analyze the role of each vision feature (\ie, CLIP~\cite{clip} and Synchformer~\cite{synchformer}) used in MMAudio~\cite{mmaudio}, which is also used in parameter initialization of our method.
Understanding how each embedding contributes to generation quality is crucial for determining how to adapt the video encoder for text conditioning.
\cref{tab:visemb} reports an ablation study by substituting each feature with its corresponding learned null embedding trained for classifier-free guidance.
Interestingly, removing the CLIP~\cite{clip} embedding actually improves the CLAP score by 0.038 while the DeSync score remains steady. 
This suggests that the CLIP embedding often introduces semantic distraction without conveying significant temporal information.
In contrast, the IB and DeSync scores deteriorate significantly when the Synchformer embedding is removed. 
This shows that the Synchformer feature contributes both semantic and temporal information for reliable audio-video alignment.

\paragraph{Input configuration}\label{sec:sync_feat}
Synchformer, consisting of audio encoder and video encoder, is learned to predict the temporal offset to evaluate audiovisual synchronization. 
In this experiment, we only use the video encoder for feature extraction, following the details of MMAudio.
Note that the architecture of the video encoder follows the Motionformer with divided space-time attention~\cite{motionformer,timesformer}.
Given an input video of 8 seconds at 25 fps, we first divide it into segments with windowing (window size of 16, hop size of 8 frames), which results in 24 segments. 
Here, a batched video data has a shape [Batch, Segments, Channel, Height, Width] by resizing $224 \times 224$ resolution without center crop. 
Each video frame is patchified and flattened in rasterized order.
After passing the video encoder, each segment results in $8$ embeddings in the temporal axis with a hidden dimension of $D=768$. The final video feature $\bv$ of a minibatch has a shape of [Batch, Segments, $t=8$, $D=768$].

\subsection{Text encoder}
To extract text embeddings, we use Flan-T5-Base~\cite{flant5}
\footnote{\url{https://huggingface.co/google/flan-t5-base}} 
to condition the video encoder.
For the audio generator, to reuse pretrained parameters from MMAudio, CLIP's text encoder~\cite{clip} is employed.

\begin{figure}[!t]
  \centering
  \includegraphics[width=0.8\columnwidth]{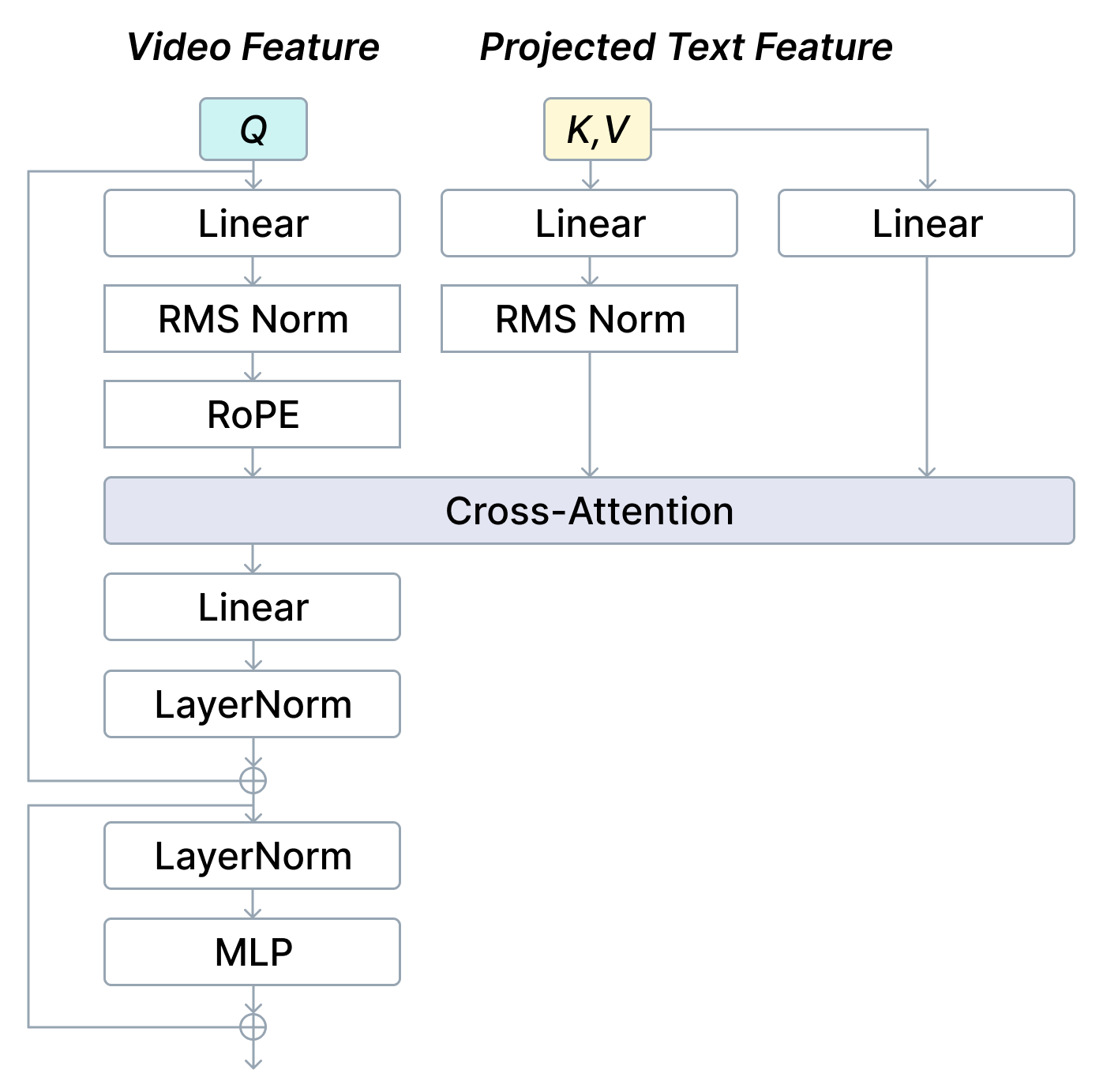}
  \caption{Detailed architecture of cross-attention used in student video encoder.}
   \label{fig:cross_attn}
\end{figure}

\begin{figure}[!t]
  \centering
  \includegraphics[width=\columnwidth]{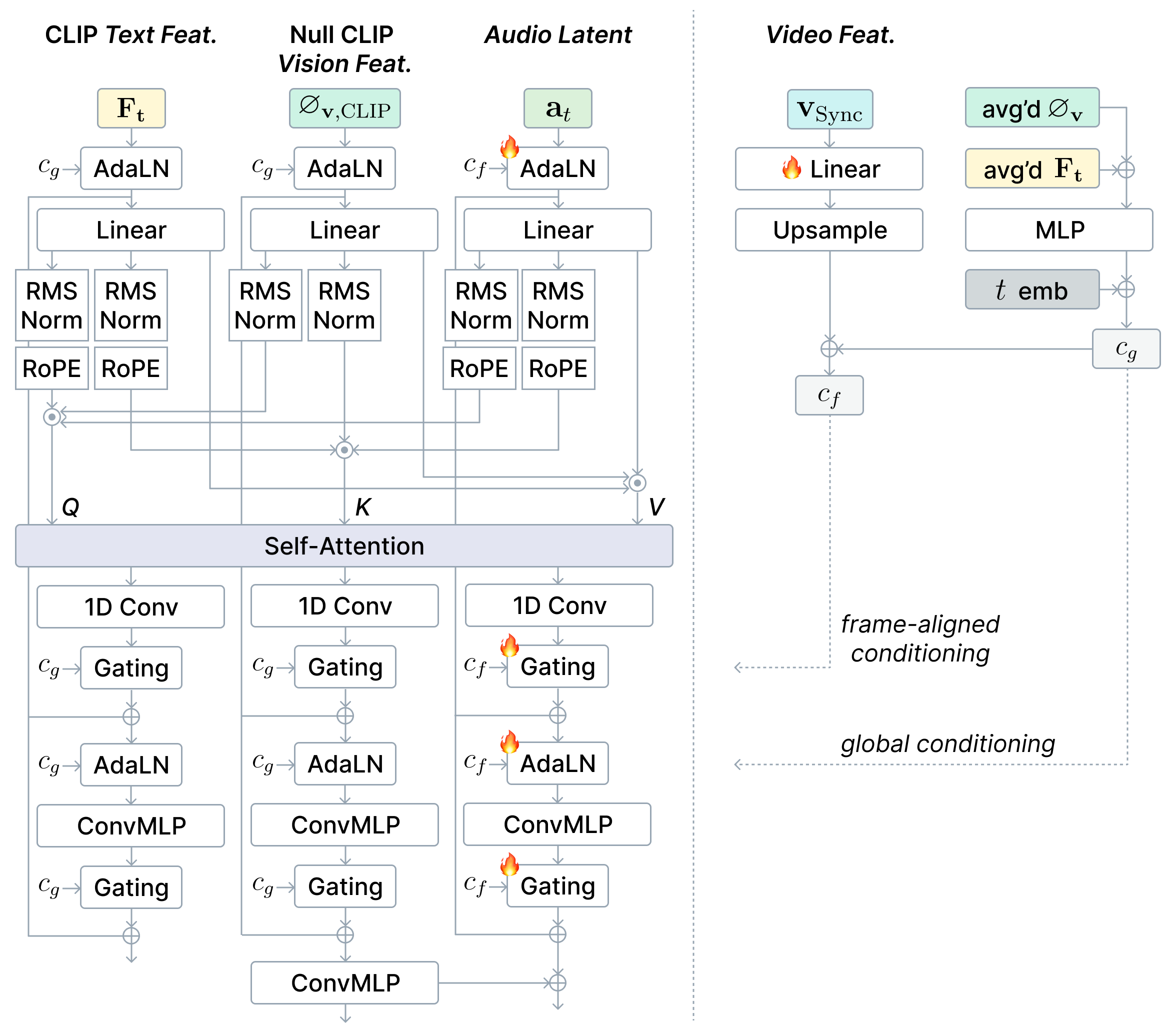}
  \caption{Detailed architecture of Multimodal transformer block in MM-DiT(Multi Modal Diffusion Transformer).}
   \label{fig:train2_detail}
\end{figure}

\subsection{Training}\label{sec:train_detail}

In the first training stage, we finetune the Synchformer~\cite{synchformer} video encoder to condition text prompts. 
We use the pretrained checkpoint \texttt{24-01-04T16-39-21} from the official implementation \footnote{\url{https://github.com/v-iashin/Synchformer}},
trained on AudioSet~\cite{audioset} using a two-stage process consisting of audio-visual contrastive learning and offset estimation.
Specifically, we train the initialized spatial attention pooling layer and a new trainable text cross-attention block, which is placed after the space-time attention blocks (see \cref{fig:cross_attn} for details). 
This process ensures parameter-efficient finetuning, updating only 14\% (19M) of the 135M total parameters.
We train for 50k steps with a batch size of 4 on a single NVIDIA RTX 4090. 
The base learning rate is set to 1e-4, with a 1k-step linear warmup schedule.

In the second training stage, we train the multimodal-conditioned audio generator.
To efficiently train the large-scale generator, we take the initial parameters from the MMAudio-small-16k model~\cite{mmaudio}.
Therefore, the architecture of the generator in \ours is similar to MMAudio.
Only 14\% (22M) of the 157M total parameters are trained in our second stage.
Concretely, we finetune the initial projection layer for the Synchformer video feature $\bv_\text{Sync}$ and all adaLN-related layers that receive the video feature as input. 
\cref{fig:train2_detail} specifies those learnable layers within a single MM-DiT~\cite{sd3,flux} block of MMAudio. 
It is worth noting that a frame-aligned conditioning $c_f$ is a function of the Synchformer video feature, while the global conditioning $c_g$ is not. 
While MMAudio originally used the CLIP image feature $\bv_\text{CLIP}$, we do not use this for conditioning by replacing with the null feature $\varnothing_\bv$.
We train for 25k steps with a batch size of 12 on a single NVIDIA RTX A6000. 
The base learning rate is set to 1e-5, also with a 1k-step linear warmup.

Common to both training stages, we utilize \texttt{bf16} automatic mixed precision (AMP). 
We employ the AdamW optimizer with $\beta_1=0.9$, $\beta_2=0.95$, and a weight decay of 1e-6, along with gradient clipping at a norm of 1. After training, we apply post-hoc EMA~\cite{posthoc_ema} with a relative width of $\sigma_\text{rel}=0.05$.

\section{\ourbench}
\label{sec:benchmark_apdx}

This section details the benchmark construction process and its resulting statistics.

\begin{figure}[t!]
  \centering
  \includegraphics[width=\columnwidth]{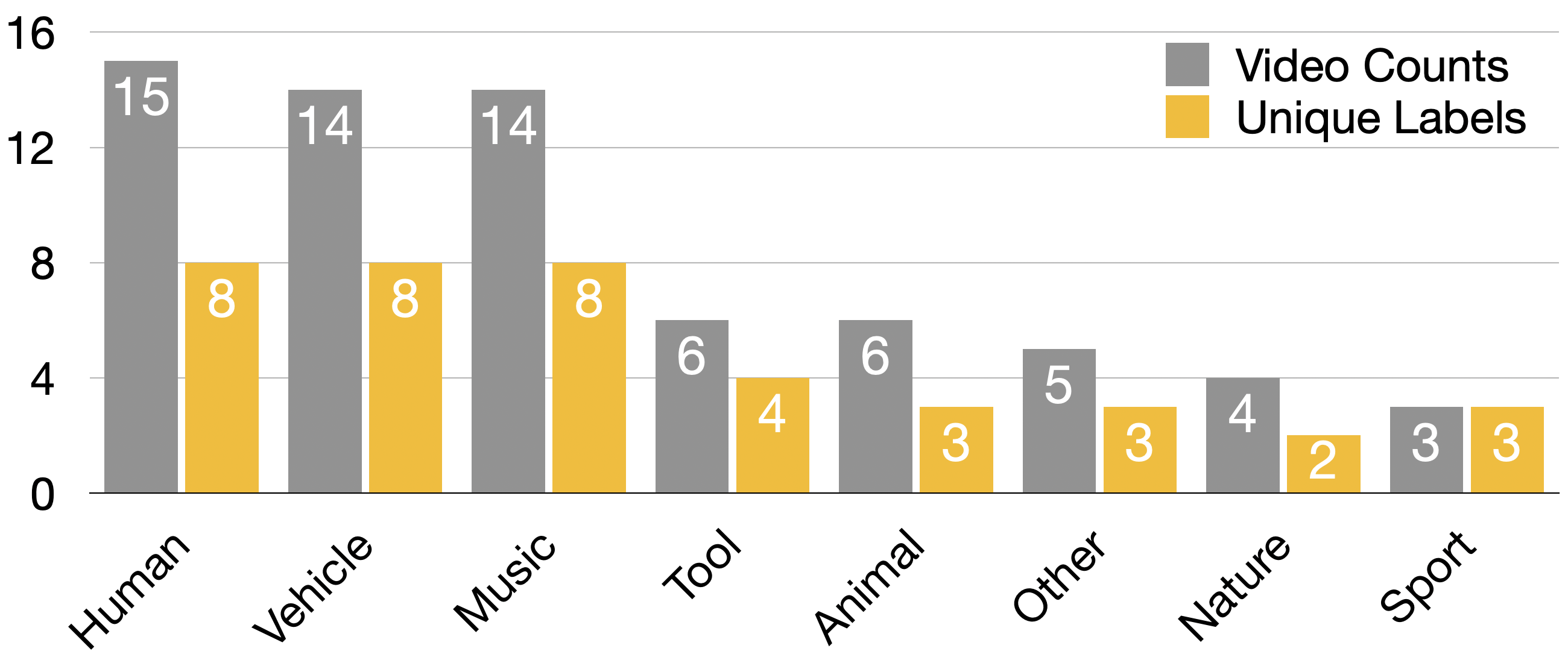}
  \caption{Statistics on single-source videos.}
   \label{fig:stat_source}
\end{figure}

\begin{figure}[t!]
    \centering
    \begin{subfigure}[]{0.5\linewidth}
        \centering 
        \includegraphics[width=\textwidth]{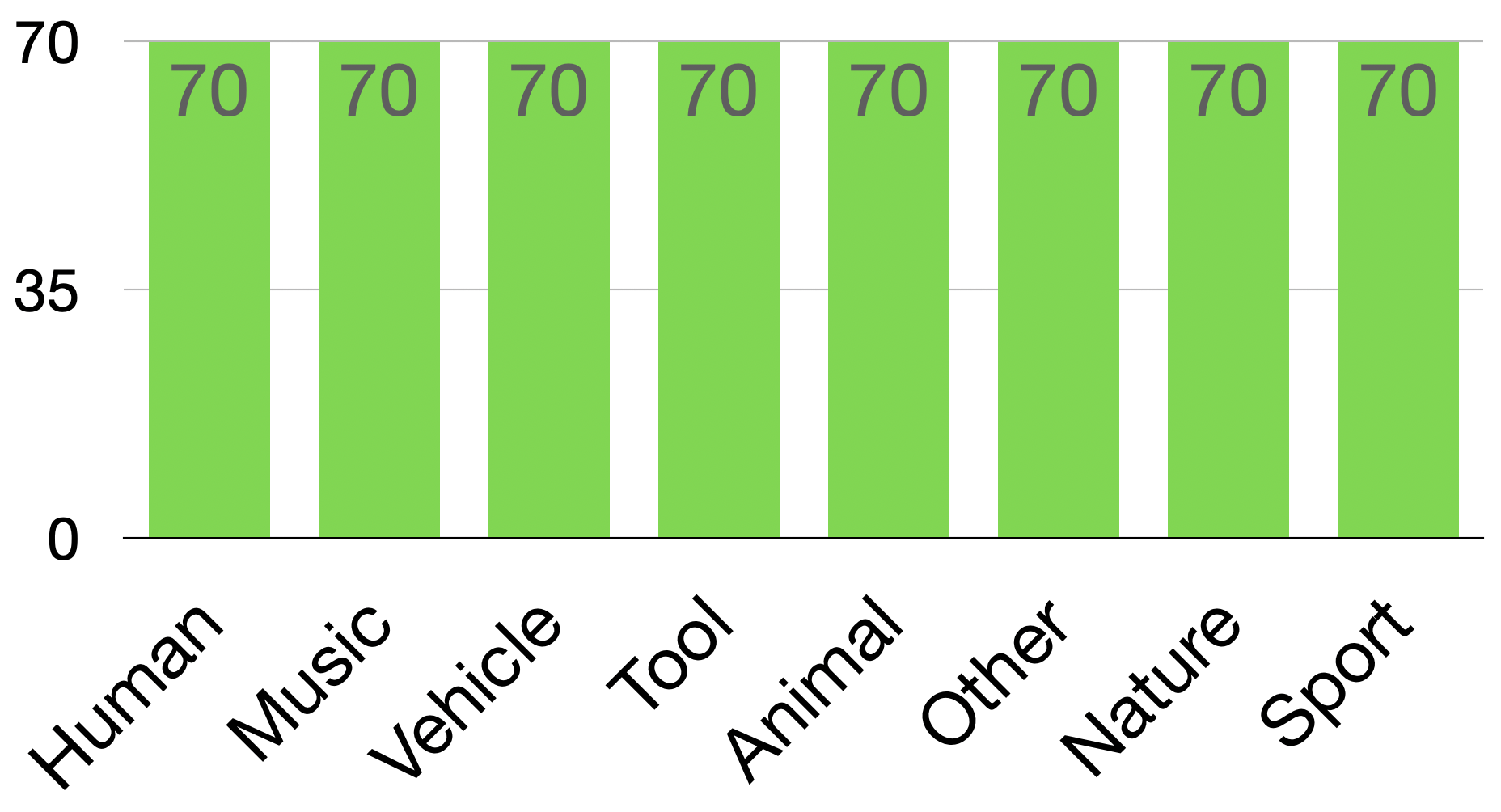}
        \caption{\textit{Inter-class}}
        \label{fig:stat_inter}
    \end{subfigure}\hfill
    \begin{subfigure}[]{0.5\linewidth}
        \centering
        \includegraphics[width=\textwidth]{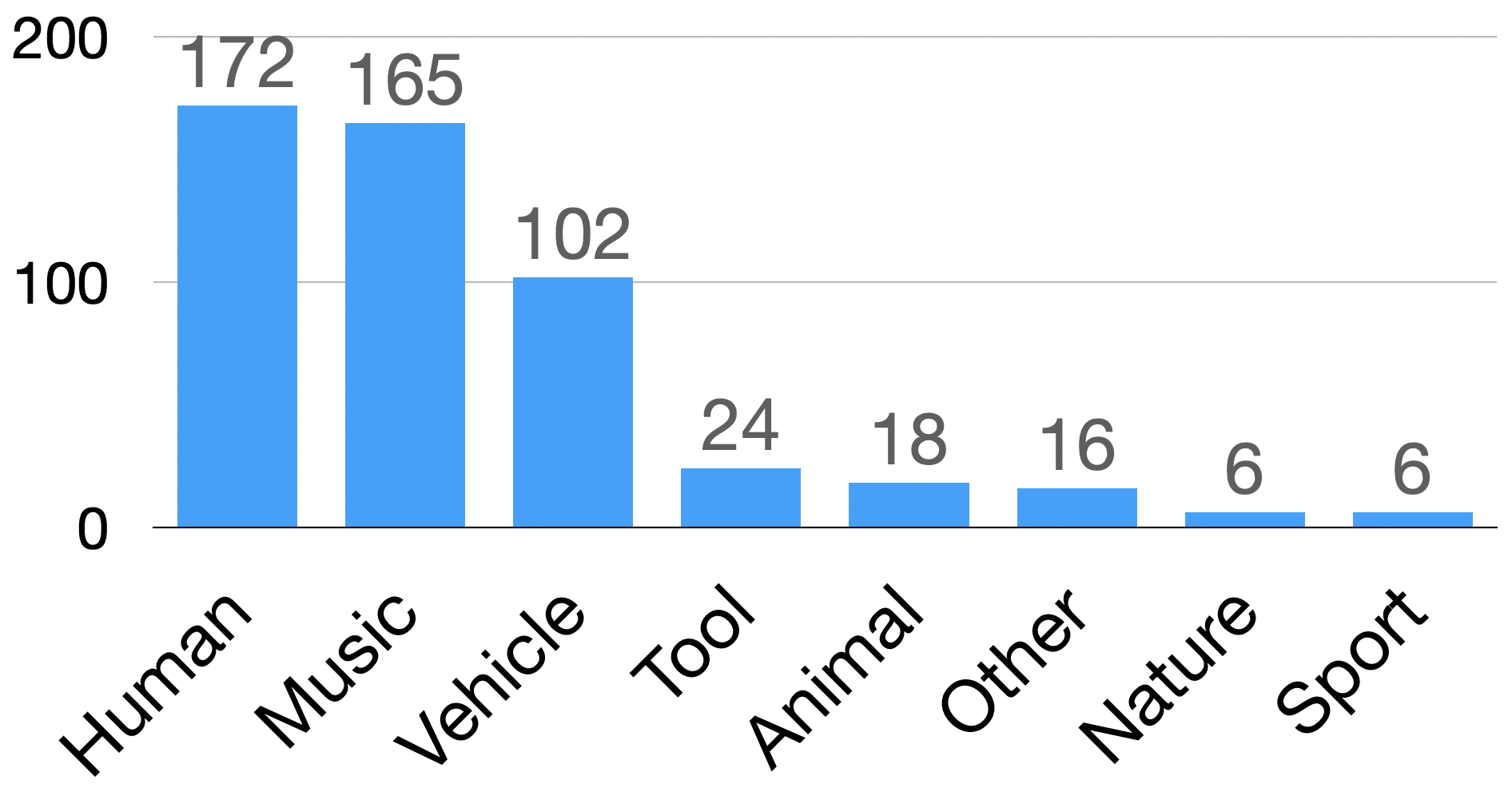}
        \caption{\textit{Intra-class}}
        \label{fig:stat_intra}
    \end{subfigure}\hfill
    \caption{Statistics on \ourbench.}
    \label{fig:stat_combined}
\end{figure}

\begin{table}[th!]
\centering
\resizebox{0.8\columnwidth}{!}{%
\begin{tabular}{@{}ll@{}}
\toprule
\textcolor{gray}{\textit{Human}} &                \\
baby crying                    & baby laughter               \\
child singing                  & male singing                \\
people burping                 & people sneezing             \\
people whispering              &     baby babbling       \\ \midrule
\textcolor{gray}{\textit{Vehicle}} &               \\
car passing by                 & driving buses               \\
driving motorcycle             & engine knocking             \\
fire truck siren               & police car siren            \\
train wheels squealing         &     airplane flyby                        \\ \midrule
\textcolor{gray}{\textit{Music}} &            \\
playing acoustic guitar        & playing banjo               \\
playing cello                  & playing electric guitar     \\
playing harmonica              & playing harp                \\
playing zither                 &     playing accordion                        \\ \midrule
\textcolor{gray}{\textit{Tool}} &            \\
lawn mowing                    & typing on computer keyboard \\
vacuum cleaner cleaning floors &     chainsawing trees                        \\ \midrule
\textcolor{gray}{\textit{Animal}} &               \\
dog barking                    & sheep bleating              \\ 
bird chirping  & \\
\midrule
\textcolor{gray}{\textit{Nature}} &          \\
waterfall burbling             &     underwater bubbling                        \\ \midrule
\textcolor{gray}{\textit{Sport}} &            \\
rope skipping                  & skateboarding               \\ 
basketball bounce & \\
\midrule
\textcolor{gray}{\textit{Other}} &          \\
firework banging               & machine gun shooting        \\ 
church bell ringing & \\
\bottomrule
\end{tabular}%
}
\vspace{-.7em}
\caption{List of 39 unique text labels in \ourbench.}
\label{tab:bench_textlist}
\end{table}

\subsection{Data collection}
As mentioned in Section \ref{ssec:exp_setup}, we acquired 67 clean, mono-source videos through automatic filtering and manual curation. These videos cover 39 unique text labels spanning 8 sound categories in \cref{tab:bench_textlist}. \cref{fig:stat_source} summarizes the category-wise statistics.
To obtain clean, mono-source audio-video-text samples, we begin with the audio-visual event annotations from UnAV-100~\cite{unav100}, a dataset containing timestamped text labels for 100 sound categories across 10,000 videos. First, we identified 907 videos that are common to both UnAV-100 and the VGGSound testset. An automatic filtering step then removed clips annotated with more than one unique sound event. Subsequently, we performed a manual verification process, retaining clips that met three strict criteria: (1) the video contains a single audible sound source with minimal background noise or off-screen sound; (2) the sounding object is clearly visible; and (3) the text annotation precisely matches the sound event. When text annotations from UnAV-100 and VGGSound differed, we manually selected the more appropriate one.

\subsection{Statistic}
An exhaustive pairing of these 67 source videos yields a total pool of 3,750 potential inter-class pairs and 560 potential intra-class pairs.
First, to construct the \textit{Inter-class} \ourbench, we sample 560 pairs from the 3,750 available, ensuring balance by target sound category as shown in \cref{fig:stat_inter}. 
We ensure a balanced selection of sample pairs across sound categories. When a category does not contain enough valid pairs, we fill the remaining slots by randomly sampling from the available pairs in that category. 
This results in the final test set of 560 inter-class pairs.
For the \textit{Intra-class} \ourbench, we manually filter the initial 560 candidate pairs to prevent semantic overlap. 
This step removes pairs where the target text prompt semantically subsumes the paired video's prompt.
For instance, a pair with the target \textit{`people whispering'} and the non-target \textit{`baby mumbling'} would be removed, as the target label could also refer to the non-target video. 
This curation process results in a final set of 511 intra-class pairs, as summarized in \cref{fig:stat_intra}.

\subsection{Pre-processing}
The target frame is randomly placed on either the left or right side of the video. All videos are processed to a $1280 \times 720$ resolution, with video encoded using the H.264 codec and audio using the AAC codec. Each video clip is 8 seconds long, with a 25 fps and an audio sample rate of 16kHz.



\section{Detailed Evaluation Setup}
\subsection{Baseline models}
\label{sec:baseline}
\paragraph{ReWaS~\cite{rewas}}
ReWaS leverages a pretrained text-to-audio (TTA) model as its generator for text-conditioned V2A. 
The model first predicts the audio's energy curve from the input video and uses this curve as a condition for the TTA model. Since ReWaS natively generates 5-second audio, we adapt it for 8-second videos by splitting each video into two overlapping 5-second segments (0-5s and 3-8s). We generate audio for each segment independently and then construct the final 8-second track by merging the first 4 seconds of the first clip (0-4s) with the last 4 seconds of the second clip (4-8s). We use the official implementation\footnote{\url{https://github.com/naver-ai/rewas}} with default parameters.

\paragraph{VinTAGe~\cite{vintage}}
VinTAGe is also a text-conditioned V2A model that aims to generate both on-screen and off-screen sounds that are semantically consistent with the text and video. As the model generates 10-second audio, we take the first 8 seconds for evaluation. We use the official code\footnote{\url{https://github.com/sakshamsingh1/vintage_aud_gen}} and default parameters for ODE sampling during inference.

\paragraph{VOS+MMAudio}
To implement segmentation-based models~\cite{hearyourclick,saganet} in our experimental setup, we employ the SoTA video object segmentation model, DEVA~\cite{deva}, and multimodal-conditioned audio generator model, MMAudio~\cite{mmaudio}.
Similar to \ours, this VOS-based pipeline takes a video and a text prompt as condition inputs to improve user controllability.
DEVA first predicts a segmentation mask for each frame based on the text prompt. Pixels outside this predicted mask are zeroed out to form a masked video.
Therefore, ideally, only the text-related target object is visible. 
This masked video is subsequently fed into MMAudio, along with the original text prompt, to generate the corresponding audio.
We used the official DEVA implementation \footnote{\url{https://github.com/hkchengrex/Tracking-Anything-with-DEVA}} 
with its default hyperparameters, which include leveraging  SAM~\cite{sam} for segmentation, applying semi-online temporal fusion of segmentation hypotheses, and disabling video resizing.

\subsection{Metrics}
\label{sec:metrics_apdx}
To assess overall audio quality, we adopt three different metrics. Fréchet Audio Distance (FAD)~\cite{fad} measures the Fréchet distance between Gaussian distributions fitted to audio embeddings from a reference set and a generated set. Kernel Audio Distance (KAD)~\cite{kad}, proposed as an unbiased and distribution-free alternative to FAD, also measures this set-wise embedding distance using the Maximum Mean Discrepancy (MMD) with a Gaussian RBF kernel. Inception Score (IS)~\cite{inception_score} evaluates both the quality and diversity of generated samples by calculating the KL divergence between the conditional label distribution for individual samples and the marginal distribution across all samples. 

For semantic alignment, we report KL divergence, CLAP~\cite{clap_laion}, and ImageBind~\cite{imagebind} scores. 
The Kullback-Leibler divergence (KL) measures audio semantic similarity using the audio classification distributions of the generated and ground-truth audio. 
CLAP and IB scores capture global semantic similarity using cosine distance between text-audio and video-audio pairs, respectively.

Finally, to assess audio-video temporal alignment, we report the DeSync score~\cite{mmaudio}, which is defined as the average predicted offset (in seconds) between the audio and video predicted by a pretrained Synchformer~\cite{synchformer}.

We use a pretrained PANNs~\cite{panns} model to extract audio embeddings for FAD, KAD, IS, and KL, as the model's features have shown a high correlation with human perception of audio quality~\cite{correlation_fad,kad}.
All metrics were calculated using open-source toolkits, including \texttt{av-benchmark} \footnote{\url{https://github.com/hkchengrex/av-benchmark}} 
and \texttt{kadtk} \footnote{\url{https://github.com/YoonjinXD/kadtk}}.

\begin{figure}[t!]
    \centering
    \includegraphics[width=\columnwidth]{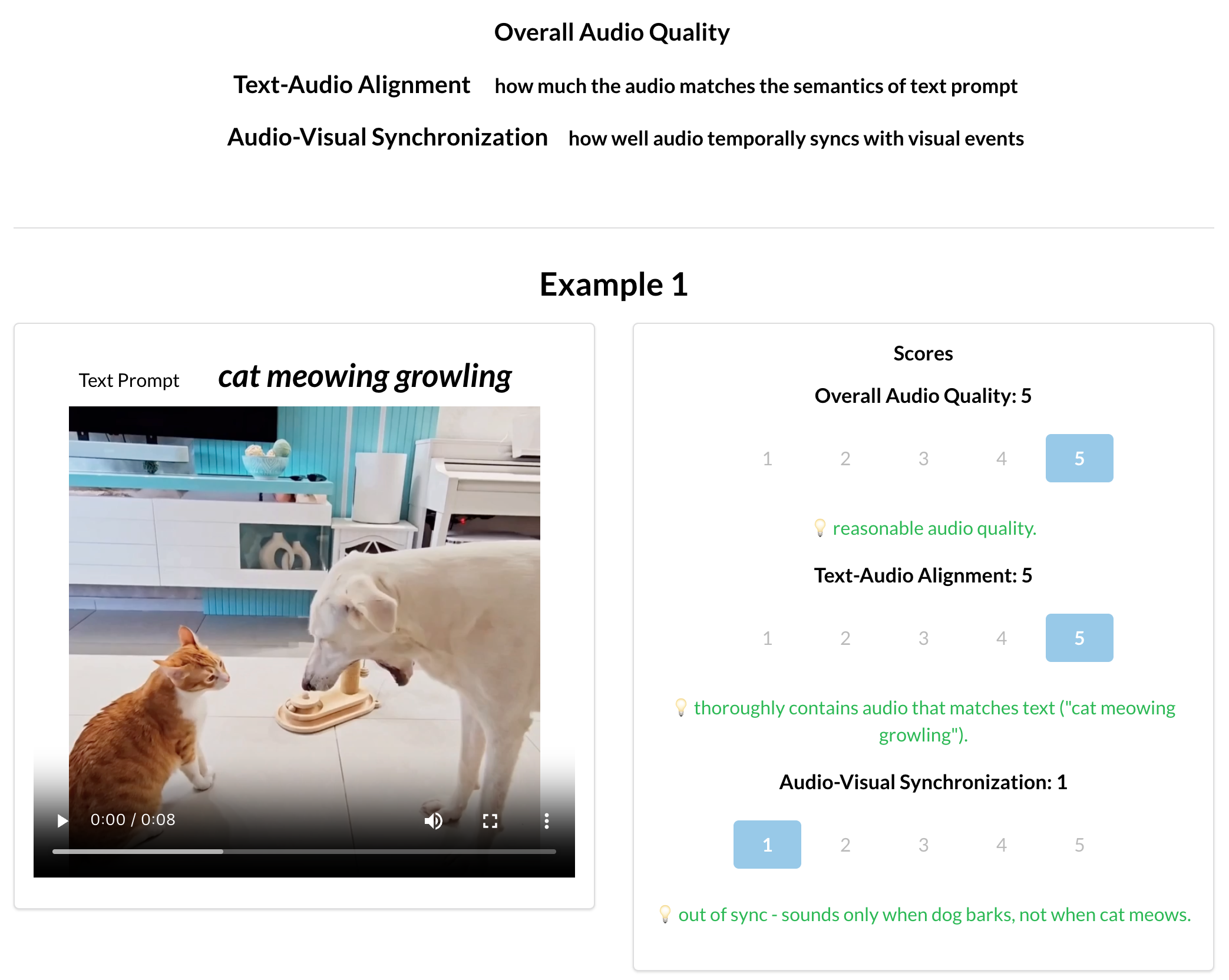}
    \caption{Tutorial example for human study to guide participants in rating audio-video-text pairs along with audio quality, text–audio alignment, and audio–video temporal alignment.}
    \label{fig:tutorial}
\end{figure}
\begin{figure}[t!]
    \centering
    \includegraphics[width=\columnwidth]{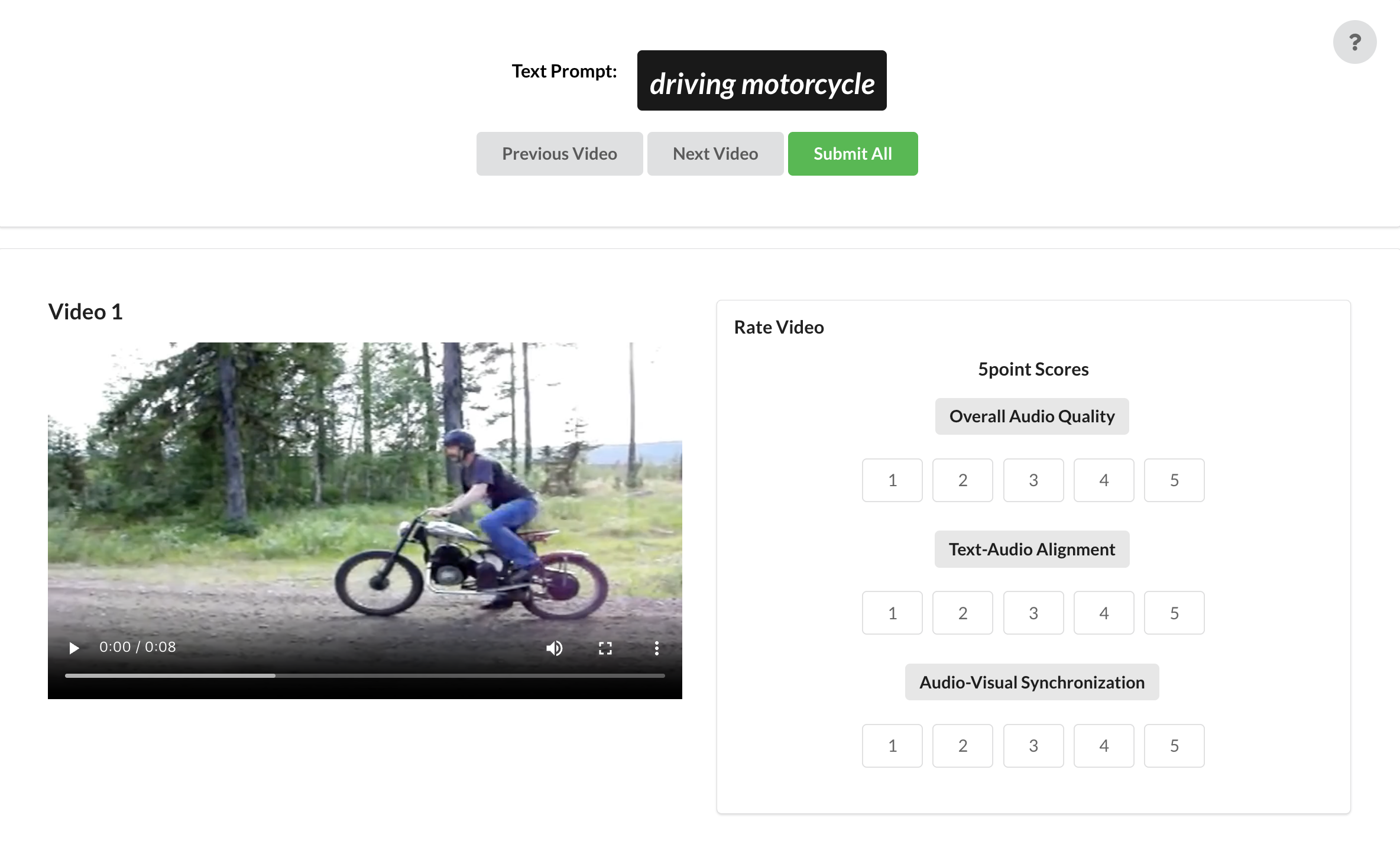}
    \caption{Web interface for human study, where participants rated audio–video-text pairs on three criteria.}
    \label{fig:human_study}
\end{figure}

\subsection{Human study}
In \cref{sec:human_study}, we conducted a human study to evaluate different text-conditioned V2A models. We provided a tutorial for each criterion (\ie, AQ, TA, VA) with 4 examples, as shown in \cref{fig:tutorial}. After watching each video clip, the participants were asked to score each criterion on a 5-point Likert scale, as shown in \cref{fig:human_study}.

\subsection{Attention visualization}
\label{sec:attn_apdx}
\cref{fig:attn_eos} visualizes the attention scores associated with the \texttt{[eos]} text token embedding by combining two attention maps: the text-guided cross-attention (with \texttt{[eos]} as the key), and the spatial-pooling map.
Both are averaged over their respective attention heads, and multiplied element-wise. 
This final visualization reveals how much the text semantics contributed to the video feature at a specific time frame.

\section{Additional Results
}

\begin{table*}[t!]
\centering
\small
\begin{tabular}{rccccccc}
\toprule
 \multirow{2}[3]{*}{\# of \reg }&
  \multicolumn{3}{c}{\cellcolor{backGray}Audio Quality} &
  \multicolumn{3}{c}{\cellcolor{backBlue}Semantic Alignment} &
  {\cellcolor{backRed}Temporal Alignment} \\
 \cmidrule(l{1pt}r{1pt}){2-4}
 \cmidrule(l{1pt}r{1pt}){5-7}
 \cmidrule(l{1pt}r{1pt}){8-8}
 &
  FAD$\downarrow$ &
  KAD$\downarrow$ &
  IS$\uparrow$ &
  KL$\downarrow$ &
  CLAP$\uparrow$ &
  IB$\uparrow$ &
  DeSync$\downarrow$ \\ \midrule
\textcolor{gray}{\textit{Inter-class}} \\
0 & 51.4 & \textbf{0.637} & 12.95 & \textbf{1.79} & 0.289 & 0.3272 & 0.756 \\
1 & \textbf{51.2} & 0.655 & 12.97 & 1.82 & 0.290 & 0.3263 & 0.760 \\
3 & \textbf{51.2} & 0.638 & 12.94 & 1.81 & \textbf{0.292} & \textbf{0.3289} & 0.745 \\
\ours / 5  & 51.7 & 0.676 & \textbf{13.07} & 1.85 & \textbf{0.292} & 0.3251 & \textbf{0.721} \\
7 & 51.9 & 0.659 & 13.02 & 1.84 & 0.289 & 0.3233 & 0.759 \\
\textcolor{gray}{\textit{Intra-class}} \\ 
0 & \textbf{36.3} & 0.485 & 9.74 & \textbf{1.01} & 0.281 & 0.3277 & 0.676 \\
1 & 37.4 & 0.510 & \textbf{9.78} & 1.03 & \textbf{0.284} & 0.3296 & 0.675 \\
3 & 36.5 & \textbf{0.474} & 9.72 & 1.03 & 0.282 & 0.3255 & 0.683 \\
\ours / 5 & 37.0 & 0.492 & 9.62 & 1.04 & 0.280 & 0.3262 & \textbf{0.639} \\ 
7 & 37.2 & 0.495 & 9.65 & 1.03 & 0.280 & \textbf{0.3300} & 0.675 \\ 
\bottomrule
\end{tabular}%
\caption{Ablation on the number of \reg tokens. Since DeSync has been dramatically changed in this ablation, we adopt five tokens as the default configuration.}
\label{tab:ablation_token}
\end{table*}

\subsection{The number of \reg tokens}\label{sec:ablation_apdx}
\cref{tab:ablation_token} shows the ablation result of all objective metrics on different numbers of \reg tokens.

\subsection{VGGSound test set}\label{sec:apdx_vggsoundtest}
\cref{tab:vgg_test} summarizes the performance of state-of-the-art text-conditioned V2A models on the VGGSound~\cite{vggsound} test set.
\ours achieves results comparable to MMAudio, showing improved semantic alignment but slightly lower temporal alignment.
This stems from the nature of VGGSound original test set, which is not curated for selective sound generation and often contains text-irrelevant sound events in videos.
Consequently, DeSync may favor holistic generation models (\ie, MMAudio) that reproduce these extraneous sounds over selective generation methods (\ie, \ours).
ReWaS~\cite{rewas} and VinTAGe~\cite{vintage} underperform in all aspects, particularly in temporal alignment. This is likely because they rely heavily on the text modality: ReWaS leverages a pretrained text-to-audio model, while VinTAGe is trained to generate both on-screen and off-screen sounds based on text descriptions.
Additionally, we observe that FAD follows the trend of KAD in \cref{tab:vgg_test} (dataset size: 15k), unlike in \cref{tab:objective_eval} (dataset size: 0.5k). This discrepancy arises because FAD is a biased estimator sensitive to sample size.

\begin{table*}[ht!]
\centering
\small
\begin{tabular}{lccccccc}
\toprule
 \multirow{2}[3]{*}{Model}&
  \multicolumn{3}{c}{\cellcolor{backGray}Audio Quality} &
  \multicolumn{3}{c}{\cellcolor{backBlue}Semantic Alignment} &
  {\cellcolor{backRed}Temporal Alignment} \\
 \cmidrule(l{1pt}r{1pt}){2-4}
 \cmidrule(l{1pt}r{1pt}){5-7}
 \cmidrule(l{1pt}r{1pt}){8-8}
 &
  FAD$\downarrow$ &
  KAD$\downarrow$ &
  IS$\uparrow$ &
  KL$\downarrow$ &
  CLAP$\uparrow$ &
  IB$\uparrow$ &
  DeSync$\downarrow$ \\ \midrule
ReWaS~\cite{rewas} & 19.96 & 1.626 & \hspace{0.3em} 7.66 & 2.42 & 0.182 & 0.1825 & 1.275 \\
VinTAGe~\cite{vintage} & 15.96 & 1.185 & \hspace{0.3em} 8.30 & 4.91 & 0.217 & 0.0486 & 1.263 \\
MMAudio-S-16k~\cite{mmaudio} & \hspace{0.3em} \textbf{7.85} & \textbf{0.338} & 18.95 & \textbf{1.75} & 0.235 & 0.2670 & \textbf{0.492} \\
\ours & \hspace{0.3em} 8.30 & 0.365 & \textbf{21.09} & 1.76 & \textbf{0.243} & \textbf{0.2688} & 0.541 \\
\bottomrule
\end{tabular}%
\vspace{-.7em}
\caption{Performance of state-of-the-art models on VGGSound~\cite{vggsound} test set. Even though \ours outperforms those methods on \ourbench, \ours still shows comparable performance on noisy VGGsound test samples.}
\label{tab:vgg_test}
\end{table*}

\subsection{Scaling up the model}
To analyze scalability, we evaluate larger versions of our audio generator $\mathcal{G}$ in \cref{tab:scale}. Following the model configurations of MMAudio~\cite{mmaudio}, the parameter counts for our small, medium, and large variants are 157M, 621M, and 1.03B, respectively. 
While generating audio at a higher sample rate necessitates more parameters, scaling the model generally improves overall performance. 
Note that the results of \ours-S-16k are reported for all other experiments.
The limited performance gain of \ours-L-44k likely stems from the restricted VGGSound; scaling to larger datasets could better exploit its capacity.

\begin{table*}[t!]
\setlength{\tabcolsep}{1em}
\centering
\small
\begin{tabular}{lccccccc}
\toprule
 \multirow{2}[3]{*}{Model}&
  \multicolumn{3}{c}{\cellcolor{backGray}Audio Quality} &
  \multicolumn{3}{c}{\cellcolor{backBlue}Semantic Alignment} &
  {\cellcolor{backRed}Temporal Alignment} \\
 \cmidrule(l{1pt}r{1pt}){2-4}
 \cmidrule(l{1pt}r{1pt}){5-7}
 \cmidrule(l{1pt}r{1pt}){8-8}
 &
  FAD$\downarrow$ &
  KAD$\downarrow$ &
  IS$\uparrow$ &
  KL$\downarrow$ &
  CLAP$\uparrow$ &
  IB$\uparrow$ &
  DeSync$\downarrow$ \\ \midrule
\textcolor{gray}{\textit{Inter-class}} \\
\ours-S-16k & 51.7 & 0.676 & 13.07 & 1.85 & 0.292 & 0.3251 & 0.721 \\
\ours-S-44k & 52.7 & 0.561 & 13.93 & 1.82 & 0.305 & 0.3490 & 0.739 \\ 
\ours-M-44k & 53.9 & 0.548 & 14.53 & 1.69 & 0.315 & 0.3586 & 0.695 \\
\ours-L-44k & 52.6 & 0.595 & 14.28 & 1.76 & 0.305 & 0.3517 & 0.691 \\
\midrule
\textcolor{gray}{\textit{Intra-class}} \\
\ours-S-16k & 37.0 & 0.492 & 9.62 & 1.04 & 0.280 & 0.3262 & 0.639 \\ 
\ours-S-44k & 38.3 & 0.340 & 10.22 & 1.12 & 0.297 & 0.3457 & 0.695 \\
\ours-M-44k & 39.0 & 0.360 & 10.34 & 1.05 & 0.295 & 0.3472 & 0.646 \\
\ours-L-44k & 38.5 & 0.346 & 10.31 & 1.13 & 0.294 & 0.3436 & 0.659 \\
\bottomrule
\end{tabular}%
\vspace{-.5em}
\caption{Performance of \ours with different sizes on \ourbench. All methods used text prompts corresponding to the target videos. The \textbf{best} scores are shown in bold, and the \underline{second-best} scores are underlined.}
\label{tab:scale}
\end{table*}

\subsection{More qualitative examples}\label{sec:apdx_eg}

\begin{figure}[t!]
    \centering
    \begin{subfigure}[]{0.5\linewidth}
        \centering 
        \includegraphics[width=\textwidth]{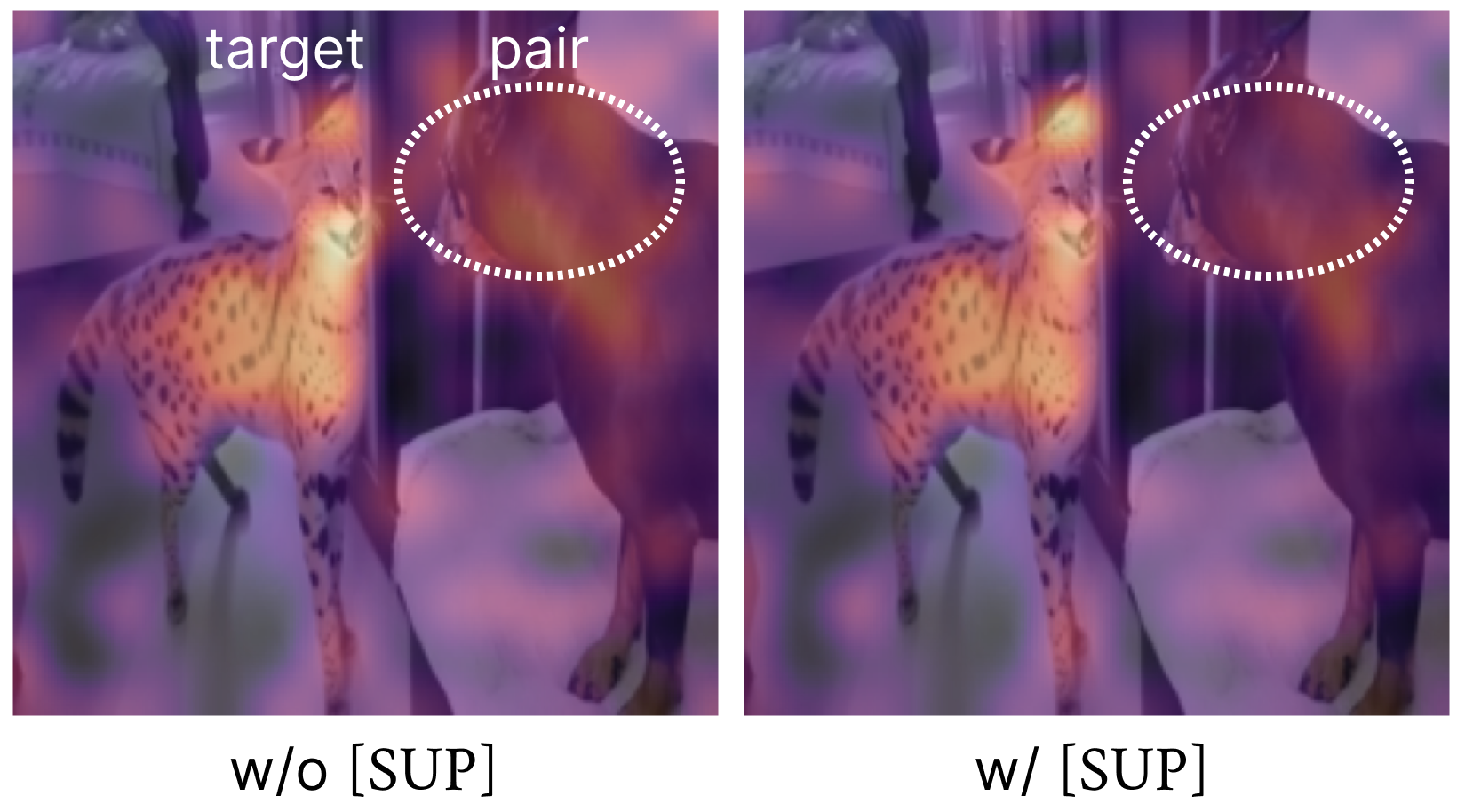}
        \caption{\textit{``Cat Meowing"}}
    \end{subfigure}\hfill
    \begin{subfigure}[]{0.5\linewidth}
        \centering
        \includegraphics[width=\textwidth]{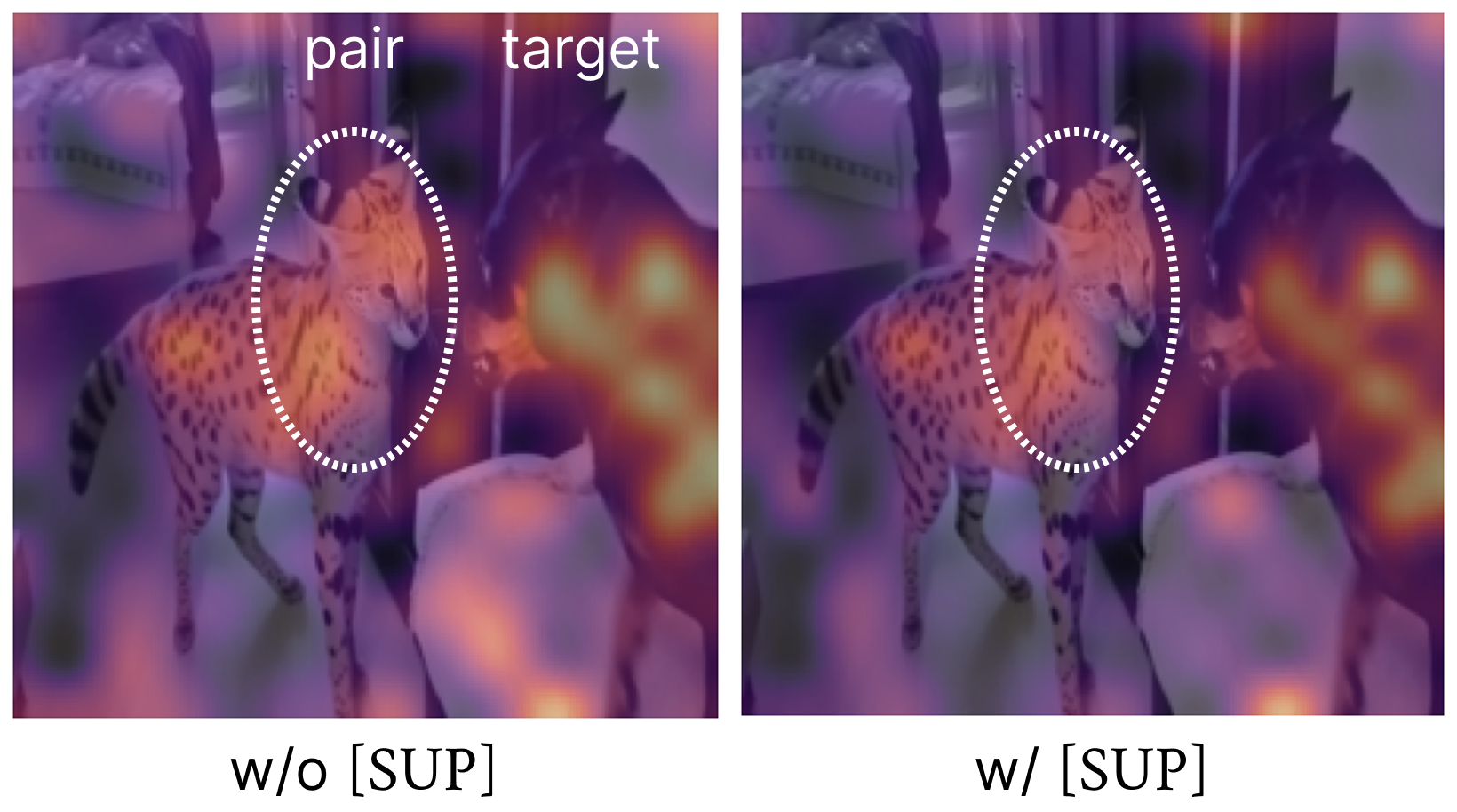}
        \caption{\textit{``Dog Barking"}}
    \end{subfigure}
    \caption{Attention visualization for \texttt{[eos]} token over real-world video frame in the last block without (left) / with (right) \reg tokens. Each subcaption denotes the corresponding target prompt.}
    \label{fig:real_world}
\end{figure}
\cref{fig:real_world} additionally provides an attention visualization of the \texttt{[eos]} token on a real-world video. By introducing our supplementary token \texttt{[SUP]}, our model focuses more effectively on the target sound source while suppressing attention toward the paired object.

\begin{figure*}[h!]
  \centering
  \def\examplewidth{0.6\textwidth}
  \begin{subfigure}[]{\examplewidth}
        \centering 
        \includegraphics[width=\linewidth]{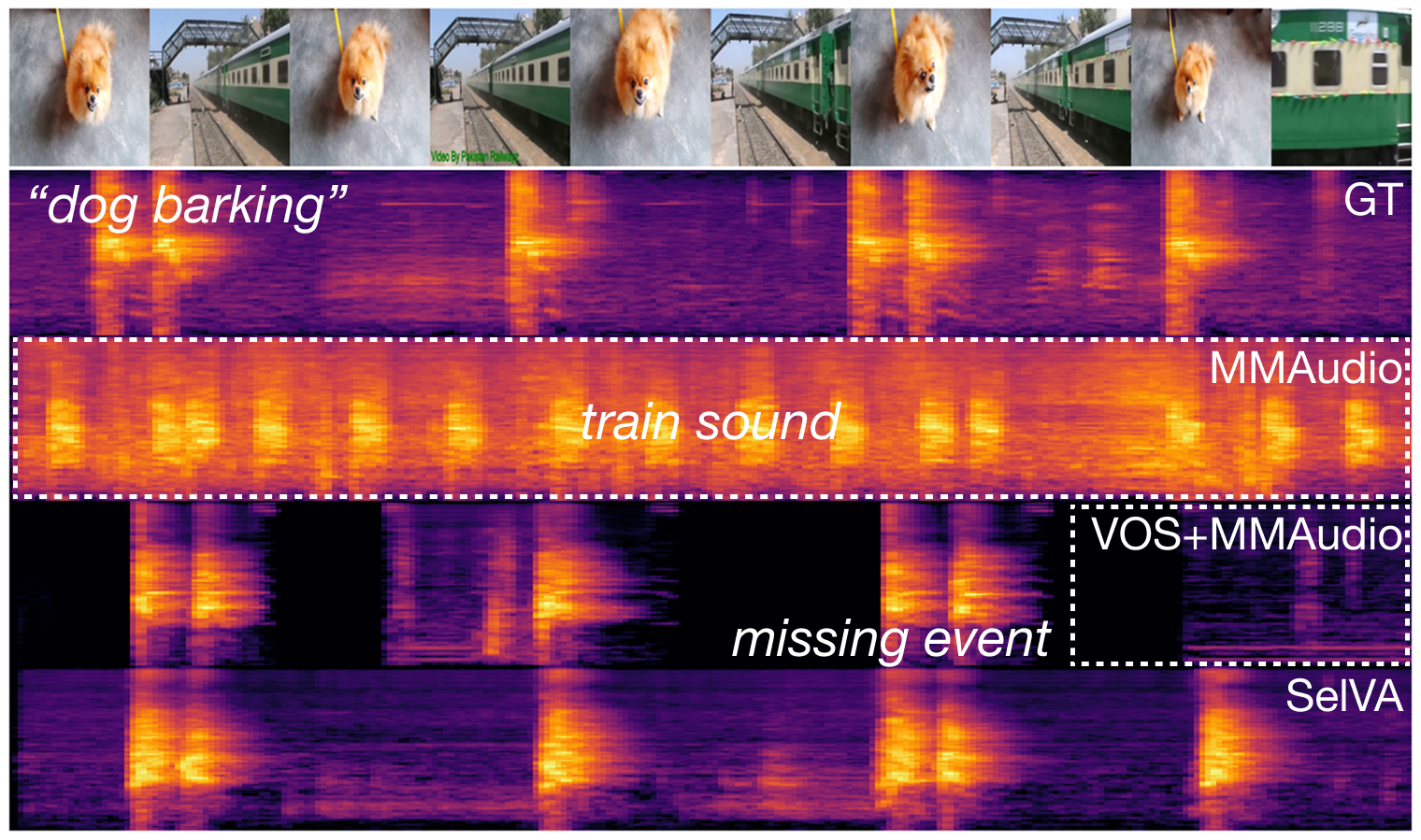}
        \caption{\textbf{\textit{``Dog barking"}} paired with \textit{``Train wheels squealing"}.}
        \label{fig:eg_dog}
    \end{subfigure}
    \begin{subfigure}[]{\examplewidth}
        \centering
        \includegraphics[width=\linewidth]{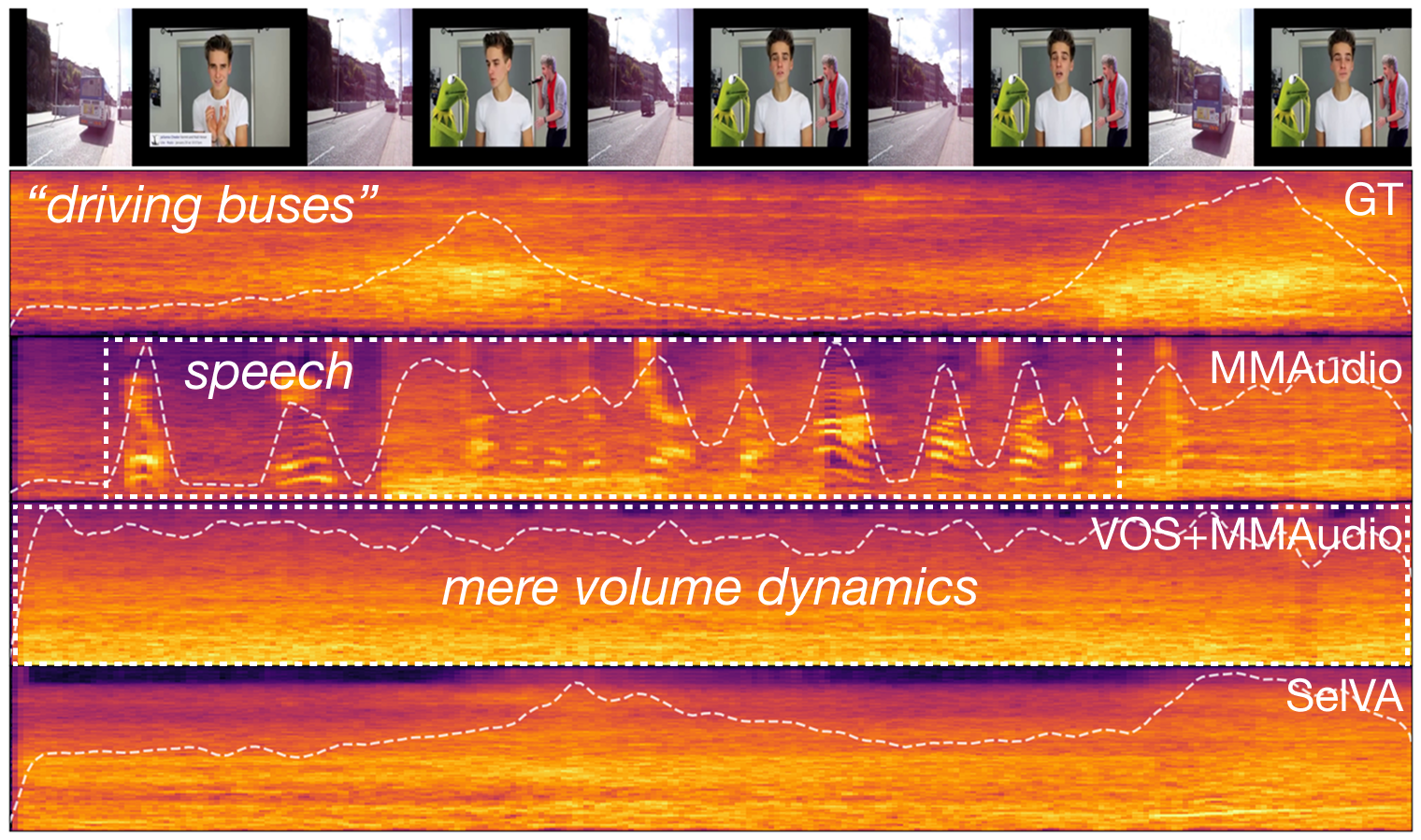}
        \caption{\textit{\textbf{``Driving buses"}} paired with \textit{``Male singing"}.}
        \label{fig:eg_bus}
    \end{subfigure}
    \begin{subfigure}[]{\examplewidth}
        \centering
        \includegraphics[width=\linewidth]{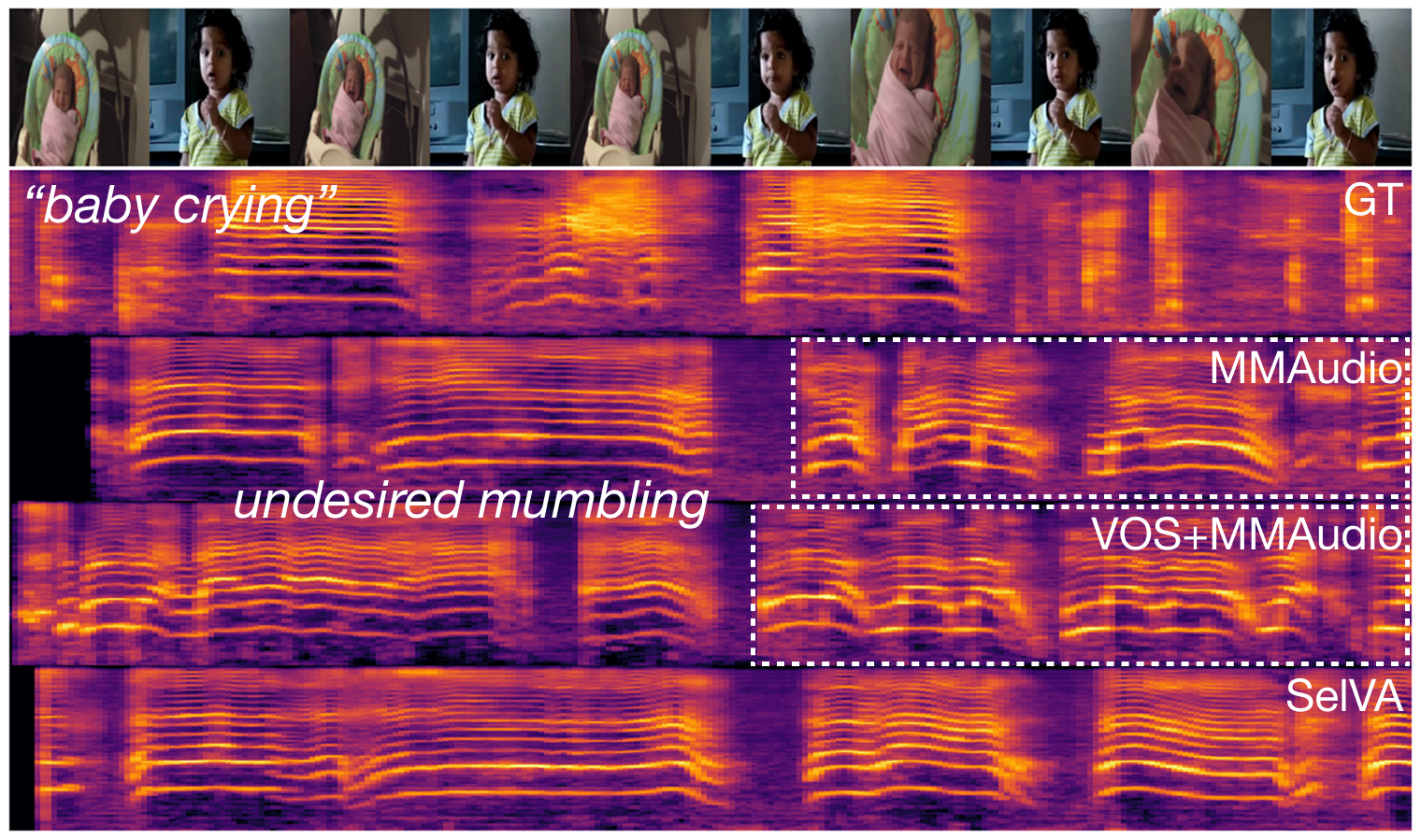}
        \caption{\textbf{\textit{``Baby crying"}} paired with \textit{``Child singing"}.}
        \label{fig:eg_baby}
    \end{subfigure}
  \caption{Qualitative performance comparison with V2A methods in \ourbench.}
   \label{fig:qual_bench}
\end{figure*}

\cref{fig:qual_bench} illustrates the mel-spectrograms of audios inferred by different models, alongside their corresponding video frames. The white dotted curve indicates the root-mean-squared (RMS) audio amplitude.

\cref{fig:eg_dog} highlights the superior selective performance of \ours. Given the target ``dog barking", MMAudio erroneously generates both the barking and a train squealing sound, with the latter correlating with the paired (non-target) video. The VOS baseline fails to capture the last barking event. In contrast, \ours faithfully generates only the dog barking sound, well-synchronized with the target video.

Example in \cref{fig:eg_bus} demonstrates the temporal synchronization capability of \ours. MMAudio again fails at selection, generating undesired male speech that stems from the paired video on the right. The VOS baseline, while correctly generating the bus sound, fails to capture its temporal dynamics (e.g., the volume change of the bus approaching and passing). We hypothesize this is due to the vision encoder's deteriorated capability; by removing the background, it loses crucial contextual information, such as the bus's size change relative to the stationary background, which implies its motion. \ours successfully captures these temporal dynamics while selectively generating the correct sound.

\cref{fig:eg_baby} showcases the semantic-level, cross-modal understanding of \ours. This intra-class example pairs a target ``baby crying" video with a ``child singing" video. The task requires the model to semantically ground the text prompt, ignoring the visually present but undesired ``child singing" event. Both MMAudio and the VOS baseline fail, generating mumble sounds synchronized with the non-target child on the right. This failure is expected for the VOS baseline, as DEVA~\cite{deva} performs object-level segmentation and cannot semantically distinguish between the two subjects based on the text. In contrast, \ours successfully leverages its text-conditioned vision encoder to generate the correct, synchronized crying sound.

\subsection{Limitation of CLAP score}\label{sec:apdx_clap}
\begin{figure*}[h!]
    \centering
    \includegraphics[width=0.6\textwidth]{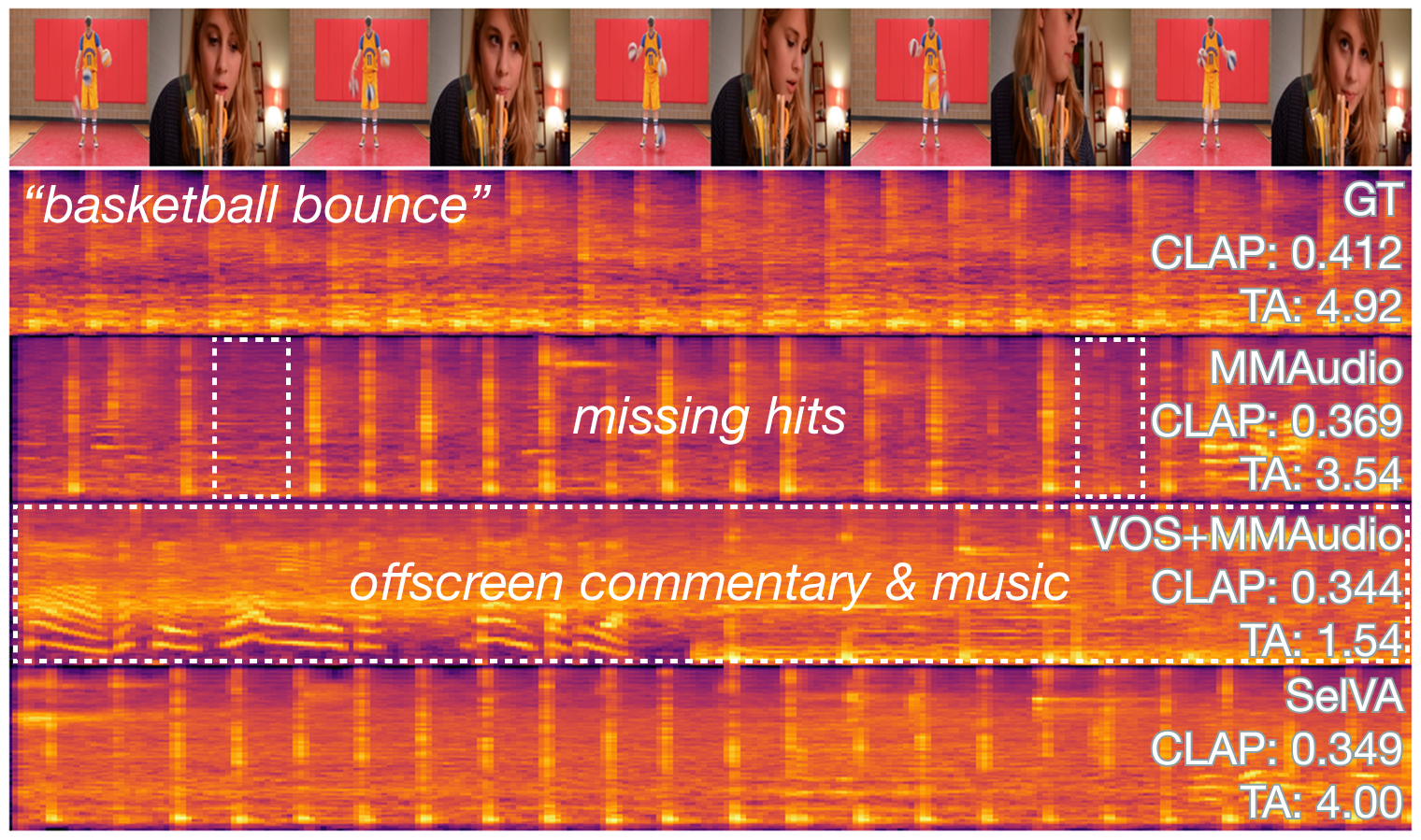}
    \caption{\textbf{\textit{``Basketball bounce"}} paired with \textit{``People whispering"}. There is a large discrepancy between the CLAP score and the human-annotated temporal alignment (TA) score.}
    \label{fig:qual_mos}
\end{figure*}

\cref{fig:qual_mos} highlights a limitation of the CLAP score~\cite{clap_laion} in capturing semantic text-audio alignment compared to human perception. 
In general, the VOS baseline was more likely to generate off-screen sounds, an error we attribute to the lack of background pixel information. 
Such artifacts may not be captured by the CLAP score.
We argue that the CLAP encoder trained on noisy audio-text pairs may not penalize the presence of such non-diegetic sounds if they occurred frequently in its training data. 
However, we found that human annotators are highly sensitive to those artifacts, in that CLAP (0.344) of VOS is comparable to that of \ours (0.349), but the temporal alignment scores (TA) differ substantially.
This result demonstrates that human study is still essential to access V2A generation methods.


\end{document}